\PassOptionsToPackage{numbers,sort&compress}{natbib}
\documentclass[Afour,sageh,times]{sagej}

\usepackage[utf8]{inputenc}
\usepackage[T1]{fontenc}

\usepackage{subcaption}
\usepackage{amsmath}

\DeclareMathOperator*{\argmin}{arg\,min}

\usepackage[ruled,vlined]{algorithm2e}
\usepackage{moreverb,url, bm, enumitem}
\usepackage{graphicx, float}
\usepackage{multirow}
\usepackage{placeins}
\usepackage{booktabs}

\usepackage[colorlinks,bookmarksopen,bookmarksnumbered,citecolor=red,urlcolor=red]{hyperref}

\newcommand\BibTeX{{\rmfamily B\kern-.05em \textsc{i\kern-.025em b}\kern-.08em
T\kern-.1667em\lower.7ex\hbox{E}\kern-.125emX}}

\def\volumeyear{202X}

\begin{document}

\runninghead{Mari\'c et al.}

\title{Diffusion-based Inverse Model of a Distributed Tactile Sensor for Object Pose Estimation}

\author{Ante Mari\'c\affilnum{1,2},
Giammarco~Caroleo\affilnum{3},
Alessandro~Albini\affilnum{3},
Julius~Jankowski\affilnum{1,2},
Perla~Maiolino\affilnum{3},
Sylvain~Calinon\affilnum{1,2}}

\affiliation{\affilnum{1}Idiap Research Institute, Martigny, Switzerland\\
\affilnum{2}École Polytechnique Fédérale de Lausanne (EPFL), Switzerland\\
\affilnum{3}Oxford Robotics Institute (ORI), University of Oxford, UK}

\corrauth{Ante Mari\'c,
Robot Learning \& Interaction group,
Idiap Research Institute,
1920 Martigny, Valais,
Switzerland.}

\email{ante.maric@idiap.ch}

\begin{abstract}
Tactile sensing provides a promising sensing modality for object pose estimation in manipulation settings where visual information is limited due to occlusion or environmental effects. However, efficiently leveraging tactile data for estimation remains a challenge due to partial observability, with single observations corresponding to multiple possible contact configurations. This limits conventional estimation approaches largely tailored to vision.
We propose to address these challenges by learning an inverse tactile sensor model using denoising diffusion. The model is conditioned on tactile observations from a distributed tactile sensor and trained in simulation using a geometric sensor model based on signed distance fields. Contact constraints are enforced during inference through single-step projection using distance and gradient information from the signed distance field. For online pose estimation, we integrate the inverse model with a particle filter through a proposal scheme that combines generated hypotheses with particles from the prior belief. Our approach is validated in simulated and real-world planar pose estimation settings, without access to visual data or tight initial pose priors. We further evaluate robustness to unmodeled contact and sensor dynamics for pose tracking in a box-pushing scenario. Compared to local sampling baselines, the inverse sensor model improves sampling efficiency and estimation accuracy while preserving multimodal beliefs across objects with varying tactile discriminability.
\end{abstract}

\keywords{
Tactile sensing, Pose estimation, Diffusion models, Particle filtering, Inverse sensor model
}

\maketitle

\begin{figure*}[h]
  \centering
  \includegraphics[width=\linewidth]{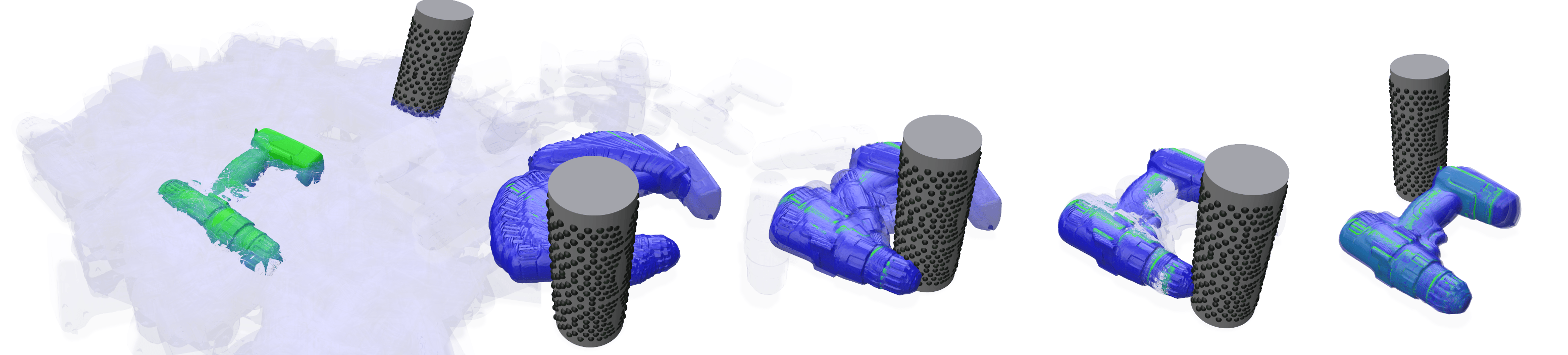}
  \caption{Particle-based belief during a simulated pose estimation experiment. Starting from a uniform prior, a cylindrical end-effector equipped with a distributed tactile sensor makes sequential contacts with a static drill. The diffusion-based inverse sensor model generates tactile-conditioned pose hypotheses at each contact, as the belief converges to the true pose. The ground-truth object is shown in green, and the belief particles in blue, with opacity proportional to their weights.}
  \label{fig:intro}
\end{figure*}

\section{Introduction}
Object pose estimation is fundamental to robot manipulation, with vision being the most commonly used sensing modality for perception. However, during physical interaction, visual observations are often unreliable due to occlusions caused by the robot itself or the surrounding environment. Tactile sensing provides a complementary modality by measuring local contact geometry and imposing geometric constraints on object pose that are independent of camera viewpoint and visual conditions. In contrast to vision, which provides global scene information, tactile observations are inherently local and reveal geometry only at regions of contact, resulting in partial observability of object pose. A single tactile observation can therefore be consistent with multiple pose hypotheses, leading to ambiguities that can persist over multiple contacts. Fully leveraging geometric constraints sensed through touch therefore requires estimation methods capable of representing and maintaining multimodal belief distributions.

Uncertainty in contact-rich manipulation tasks has long been studied in vision-free settings, with the goal of reducing pose uncertainty through sensing and planning over sequences of contacts \citep{erdmann_exploration_1988, lynch_manipulation_1992}. Within this context, model-based approaches to tactile pose estimation provide principled ways of reasoning about contact geometry and uncertainty, but are typically tailored to specific sensor configurations or interaction assumptions \citep{petrovskaya_global_2011, koval_manifold_2017}. Such approaches commonly rely on forward sensor models to evaluate observation likelihoods. In practice, however, the highly nonlinear relationship between object pose and tactile observations, together with geometric constraints imposed by contact, makes it challenging to efficiently generate contact-consistent hypotheses.
Recent learned approaches have shown promising results for tactile estimation \citep{bauza_tac2pose_2023, suresh_neural_2023}, but existing work has largely focused on high-resolution, localized sensors. Accommodating a broader range of tactile technologies remains challenging due to costly data collection and lack of general-purpose simulation tools. Moreover, many existing methods rely on tight pose priors, limiting applicability when prior information is weak or unavailable. Diffusion-based generative models \citep{ho_denoising_2020} offer a promising direction for addressing these limitations, as they can be conditioned on observations to produce diverse samples from multimodal distributions.

In this work, we propose leveraging denoising diffusion to learn an inverse tactile sensor model for hypothesis generation in vision-free object pose estimation. Specifically, we introduce a geometry-aware inverse model that generates samples from a multimodal distribution over contact configurations between the sensor and the object, conditioned on tactile observations. We consider a tactile sensor composed of a deformable array of sensing elements (taxels), that provide local contact information across parts of the robot surface~\citep{maiolino_flexible_2013}. The model learns to sample diverse contact configurations consistent with a given observation and is integrated with a particle filter for online object pose estimation without requiring tight initial pose priors.
Our main contributions are as follows:
\begin{enumerate}[label=\roman*), itemsep=0pt, topsep=2pt, leftmargin=*]
    \item A diffusion-based inverse tactile sensor model that generates multimodal pose hypotheses consistent with tactile observations and contact constraints.
    \item A tactile observation model and contact simulation framework enabling efficient generation of training data without real-world data collection.
    \item A belief-informed particle injection scheme that integrates generated hypotheses with a particle filter for online pose estimation without tight initial pose priors.
\end{enumerate}
We validate the proposed approach in a vision-free planar setting, with object pose estimation experiments demonstrating improved sample efficiency and accuracy compared to local sampling baselines. We further evaluate robustness to unmodeled sensor dynamics and sim-to-real transfer in real-world object pose estimation and tracking scenarios using a sensorized end-effector.

The remainder of this paper is organized as follows: Section~\ref{sec:related_work} reviews related work on tactile sensing and pose estimation. Section~\ref{sec:problem_formulation} introduces the problem formulation and provides an overview of our approach. Sections~\ref{sec:contact_generation} and~\ref{sec:sensor_model} describe the geometric contact model and tactile observation model used for data generation and likelihood evaluation. Section~\ref{sec:inverse_observation_model} presents the diffusion-based inverse tactile sensor model with geometry-aware inference, and Section~\ref{sec:particle_filter} describes its integration with a particle filter via belief-informed particle injection. Section~\ref{sec:experiments} presents experimental validation and results that are further discussed in Section~\ref{sec:discussion}.
\vspace{12px}

\section{Related Work}
\label{sec:related_work}
\subsection{Tactile Pose Estimation}
Early works in vision-free manipulation demonstrated that pose uncertainty can be reduced through contact interactions that exploit task mechanics and contact constraints \citep{erdmann_exploration_1988, lynch_manipulation_1992}. Such motion-based approaches continue to inform recent works that combine belief dynamics with contact-informed optimization for robust manipulation under uncertainty \citep{jankowski_robust_2024}.
Sensor-based approaches take a complementary view, decoupling estimation from planning by using tactile feedback to update state estimates. Bayesian formulations allow sequential tactile observations to be integrated over time to maintain a belief over object pose \citep{petrovskaya_global_2011}.
Due to partial observability, however, tactile estimation from force-torque and proprioceptive feedback typically requires multiple simultaneous contacts to capture underlying geometric constraints, or relies on strong priors on task structure and initial state.

Advances in tactile sensing technologies have enabled richer contact feedback for robotic manipulation, with seminal works focused on sensing microvibrations~\citep{fishel_sensing_2012}. Recent works exploit camera-based sensors \citep{yuan2017gelsight, lambeta2020digit}, which offer local high-resolution measurements of contact geometry. Settings requiring whole-body awareness, where contact may occur at multiple locations, have motivated the development of distributed tactile sensors that can cover large parts of the robot surface with arrays of sensing elements~\citep{ohmura2006conformable, maiolino_flexible_2013, cheng_comprehensive_2019,kohlbrenner_gentact_2025}. Across both local and distributed modalities, tactile information has been used to address partial observability stemming from occlusions and inform planning in manipulation tasks~\citep{kelestemur_tactile_2022, albini_exploiting_2021}. Nevertheless, limited spatial coverage and resolution remain challenging, motivating estimation methods that efficiently leverage tactile observations for added robustness.

\subsection{Bayesian Filtering and Proposal Mechanisms}
Following common methods in vision-based state estimation, tactile pose estimation approaches often adopt a Bayesian filtering framework to maintain a belief over object pose. In particular, particle filters have been adapted to tactile settings to represent multimodal pose distributions using non-parametric representations~\citep{corcoran_measurement_2010, petrovskaya_global_2011, chalon_online_2013}. Unlike vision-based or multi-sensor fusion settings, vision-free estimation requires reasoning about object pose from sparse and ambiguous contact information. The nonlinear relationship between pose and observations, together with hard contact constraints, makes it difficult to efficiently generate high-likelihood hypotheses, leaving conventional sampling methods prone to particle depletion~\citep{thrun_probabilistic_nodate}. Such methods therefore typically require strong task priors or good initial estimates, relying on Gaussian sampling around existing high-likelihood particles to maintain sufficient coverage of the posterior.

Subsequent works have explored more efficient proposal mechanisms to generate informative pose hypotheses consistent with tactile observations. For example, \citet{koval_manifold_2017} sample pose hypotheses from \textit{contact manifolds} that describe contact constraints between the robot and object geometries, with tactile measurements and dynamics models used to guide sampling toward high-likelihood regions. Contact manifolds can, however, be analytically formulated only for simple geometries, and sampling them using numerical methods such as rejection sampling is computationally expensive. \citet{wirnshofer_state_2019} incorporate object state as part of the system dynamics model to bias sampling toward physically plausible contact configurations using dynamics rollouts and pose error of compliant manipulators as observations. Such approaches successfully leverage prior domain knowledge for state estimation, but incorporating complex task dynamics and more informative tactile observations remains challenging.

\subsection{Learning-based and Generative Approaches}
An increasingly popular way of addressing the challenges of efficiently generating pose estimates under partial observability and contact constraints is through learning. Differentiable particle filters (DPFs) have been developed to address such issues in general estimation settings through end-to-end learning of different components of the particle filter~\citep{jonschkowski_differentiable_2018}. \citet{rostel_learning_2022} propose such a framework for tactile in-hand manipulation with a multi-fingered hand, relying on multiple simultaneous contacts to generate pose hypotheses from a unimodal proposal distribution. \citet{suresh2022midastouch} utilize a learned tactile code network to localize a vision-based tactile sensor over sliding contact using a particle filter. Similarly, \citet{bauza_tac2pose_2023} learn object-specific embeddings that associate tactile observations with simulated contact configurations. Learned observation models of visuotactile sensors~\citep{sodhi_leo_2022} have also been used as part of smoothing approaches using factor graphs for pose estimation and tracking of sliding objects. \citet{sodhi_patchgraph_2022} learn surface normals for in-hand tracking of unknown objects using visuotactile sensors in a factor graph framework. Similarly, \citet{zhao_fingerslam_2023} combine visuotactile feedback with vision to simultaneously localize and reconstruct objects during in-hand manipulation. Other works learn implicit neural representations for tactile reconstruction of object shape and pose~\citep{suresh_neural_2023}.
A common assumption across these methods is the availability of at least one strong source of geometric information, such as high-resolution tactile imprints from localized vision-based sensors, camera observations, or multiple simultaneous contacts, which provide informative geometric constraints and strong pose priors. Vision-free tactile estimation settings without tight initial priors or multiple simultaneous contacts remain challenging as they require reasoning over multimodal pose distributions from sparse and ambiguous tactile data.

Score-based generative models based on diffusion and flow matching~\citep{ho_denoising_2020, lipman2023flow} have been widely applied in computer vision to address partial observability caused by occlusions in pose estimation from RGB images~\citep{xu_6d-diff_2024} or partial point clouds~\citep{moller_particle-based_2024}. In robotics, diffusion models have demonstrated strong performance for policy learning~\citep{chi_diffusion_2025}, and have been formulated in $SE(3)$ for more robust learning in 6-DoF manipulation tasks~\citep{urain_se3-diffusionfields_2023}. Recent work by \citet{jin_se3-poseflow_2025} proposed a framework using flow matching on the $SE(3)$ manifold for object pose estimation from RGB-D data, aimed at manipulation under partial observability.
Similar to these works, we learn to sample pose hypotheses from multimodal distributions using a denoising diffusion implicit model (DDIM)~\citep{song_denoising_2022}. Specifically, we learn an inverse sensor model of a distributed tactile sensor~\citep{maiolino_flexible_2013} for object pose estimation in a vision-free setting. The inverse model implicitly captures the contact manifold~\citep{koval_manifold_2017} while relying on a lightweight denoiser architecture and geometric projection for exact constraint satisfaction. We condition the DDIM on single observations from a distributed sensor and integrate it with a particle filter for tracking pose beliefs over sequential contacts.

A practical challenge for learning generative tactile models is training data availability. While visuotactile sensors benefit from high-fidelity simulators~\citep{wang_tacto_2022}, distributed tactile skins lack comparable tools due to computational cost and diverse transduction principles~\citep{albini_representing_2025}. We therefore adopt a sensor output model based on contact distances, which enables scalable and computationally efficient generation of simulated training data.

\begin{figure*}[t]
  \centering
  \includegraphics[width=\linewidth]{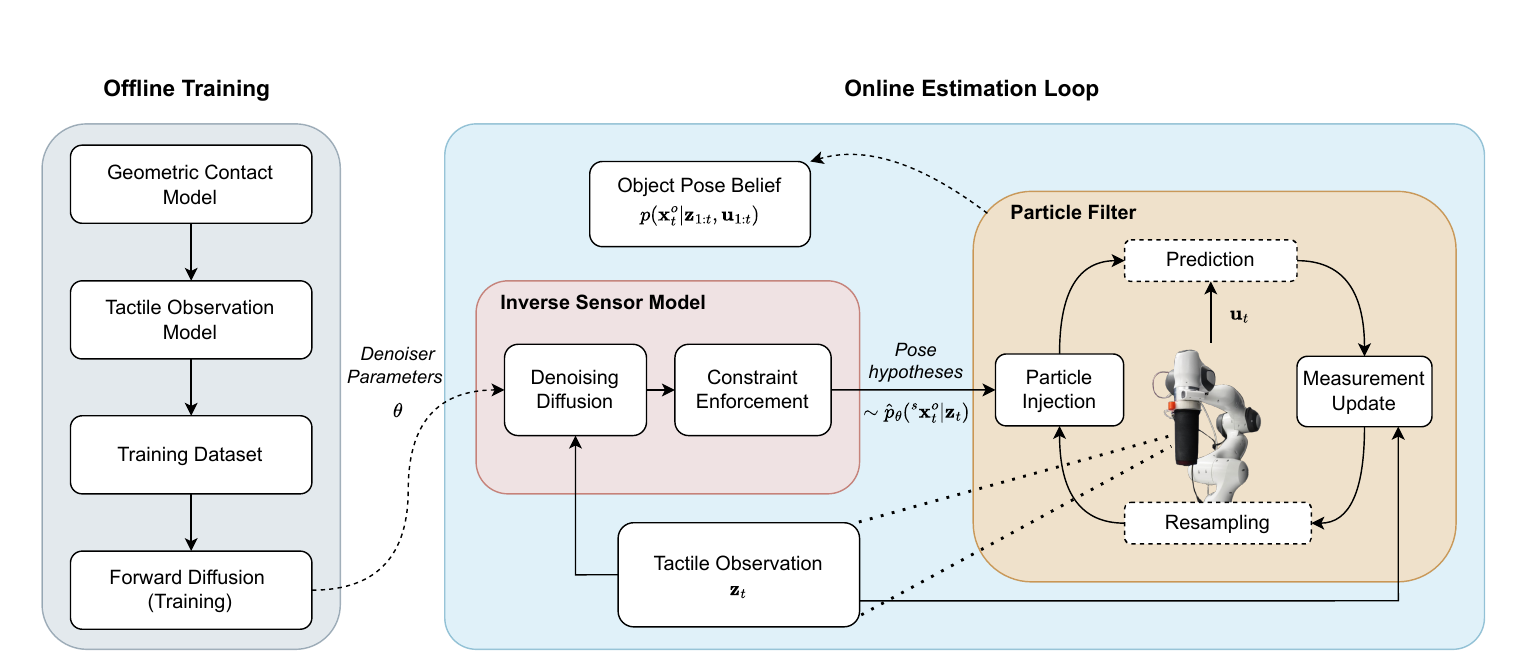}
  \caption{Overview of the proposed framework. Offline, a denoising diffusion implicit model (DDIM) is trained on tactile observations from a simulated contact dataset. Online, the trained model generates pose hypotheses conditioned on incoming tactile observations from a sensorized end-effector, with SDF-based projection enforcing contact constraints. Generated hypotheses are injected as proposal particles in a particle filter that updates a belief over object pose across sequential contacts.}
  \label{fig:framework}
\end{figure*}
\section{Problem Formulation \& Approach}
\label{sec:problem_formulation}

We consider a setting in which a robot manipulator is equipped with a distributed tactile sensor covering part of its surface. As the robot operates in the workspace, our goal is to estimate the pose of an object with known geometry through incoming tactile observations. We assume no additional exteroceptive sensing modalities, such as vision, and no tight initial priors on object pose. Based on a sequence of known robot states $\mathbf{u}_{1:t}$ and corresponding tactile observations $\mathbf{z}_{1:t}$, we track a belief over the object pose $\mathbf{x}^o_t\in \mathcal{X}^o$ at time $t$ defined as
\begin{align}
    \label{eq:belief}
    \text{bel}(\mathbf{x}^o_t) = p(\mathbf{x}^o_t \mid \mathbf{z}_{1:t}, \mathbf{u}_{1:t}),
\end{align}
starting from a uniform prior $\text{bel}(\mathbf{x}^o_0) = \mathcal{U}(\mathcal{X}^o)$ over a bounded state space. As visualized in Figure~\ref{fig:intro}, partial observability can lead to multimodal belief distributions that require progressive refinement over multiple contacts.

\paragraph{Assumptions.}
We consider a planar manipulation setting and define the object state $\mathbf{x}^o_t$ at time $t$ as a pose in $\mathcal{X}^o \subset SE(2)$, expressed as $\mathbf{x}^o_t = [q_x, q_y, \theta]^\top$, where $q_x, q_y$ represent planar position coordinates, and $\theta$ the object rotation angle around the $z$-axis. Object geometry is assumed to be known. We further assume that the robot is equipped with a cylindrical end-effector covered by a distributed tactile array of $N_{\text{tax}}$ taxels, providing observations $\mathbf{z}_t\in\mathbb{R}^{N_{\text{tax}}}$. The end-effector (sensor), visualized in Figure~\ref{fig:setup}, moves at a fixed height with its axis perpendicular to the ground plane, constraining the robot task space to $\mathcal{X}^s \subset SE(2)$. We assume that the robot has sufficiently high stiffness such that contact forces do not significantly affect its configuration, allowing us to treat commanded end-effector pose $\mathbf{u}_t \in \mathcal{X}^s$ as the true robot state.

\paragraph{Approach.}
\label{sec:method_overview}
To address partial observability in tactile pose estimation, we propose a modular framework combining a learned inverse sensor model with analytical constraint enforcement and Bayesian filtering. Since contact constraints do not need to be perfectly captured by the learned model, this approach enables faster training and inference, while allowing integration with existing dynamics and observation models.
The inverse sensor model is trained in simulation to generate multimodal object pose hypotheses conditioned on single tactile observations. The online estimation framework, illustrated in Figure~\ref{fig:framework}, proceeds as follows: given a tactile observation, the inverse sensor model generates multiple pose hypotheses, and geometric constraints are enforced analytically at inference time. These hypotheses are then integrated into a particle filter that maintains a belief distribution over object pose across sequential contacts.

The following sections detail each component: contact modeling and geometric constraints (Section~\ref{sec:contact_generation}), the tactile observation model (Section~\ref{sec:sensor_model}), the diffusion-based inverse sensor model (Section~\ref{sec:inverse_observation_model}), and the particle filter with belief-informed particle injection (Section~\ref{sec:particle_filter}).

\section{Geometric Contact Modeling}
\label{sec:contact_generation}
Valid contact configurations lie on a lower-dimensional manifold defined by geometric constraints between the sensor and object surfaces \citep{koval_manifold_2017}. We enforce these constraints analytically using distance and gradient information from a signed distance field. This serves two main roles: \textit{(i)} generating training data with valid contact configurations, and \textit{(ii)} enforcing contact constraints in generated pose hypotheses at inference time. 

\subsection{SDF-based Contact Constraints}
We represent object geometry using a signed distance field (SDF). For an object with interior $\Omega \subset \mathbb{R}^3$ and smooth boundary $\partial\Omega$, the SDF $\phi : \mathbb{R}^3 \to \mathbb{R}$ is defined as
\begin{equation}
    \label{eq:sdf}
\phi(\mathbf{q}) = \begin{cases}
d(\mathbf{q}, \partial\Omega) & \text{if } \mathbf{q} \in \Omega^c \text{ (exterior)}, \\
0 & \text{if } \mathbf{q} \in \partial\Omega \text{ (boundary)}, \\
-d(\mathbf{q}, \partial\Omega) & \text{if } \mathbf{q} \in \Omega \text{ (interior)},
\end{cases}
\end{equation}
where the unsigned Euclidean distance from point $\mathbf{q}\in \mathbb{R}^3$ to the object surface is
\begin{equation}
d(\mathbf{q}, \partial\Omega) = \inf_{\mathbf{q}_s \in \partial\Omega} \|\mathbf{q} - \mathbf{q}_s\|.
\end{equation}
We assume that $\phi$ is a smooth SDF approximation that is differentiable almost everywhere. It can be formulated using numerical precomputation and interpolation, or through learned representations, enabling fast queries of distance values and gradients for any point within a bounding box. This allows for fast geometric computations and gradient-based projection for constraint enforcement and sensor modeling. An example SDF representation is visualized in Figure~\ref{fig:sensor_model}.

For an object represented by SDF $\phi$, we consider a pair of object and end-effector poses $(\mathbf{x}^o, \mathbf{u})$ to constitute a valid contact configuration if
\begin{equation}
    \label{eq:contact_constraint}
\delta_{\text{pen}} < \phi({}^o\mathbf{q}) < \delta_{\text{max}}
\end{equation}
holds for every point $\mathbf{q}$ on the sensor surface. In the above, ${}^o\mathbf{q}$ denotes that $\mathbf{q}$ is expressed in the object frame. Signed distance thresholds $\delta_{\text{pen}}$ and $\delta_{\text{max}}$ control the softness of the contact constraints. Setting $\delta_{\text{pen}} < 0$ allows for small penetration depths to approximate the compliant layers of a soft tactile sensor, as visualized in Figure~\ref{fig:setup}.

\subsection{Gradient-based Projection}
For a given initial pair of poses $(\mathbf{x}^o, \mathbf{u})$, contact constraints are enforced using SDF-based projection. We form the projection vector $\Delta\mathbf{q}$ by finding the point on the sensor axis closest to the object surface. Specifically, we sample points $\{{}^o\mathbf{q}^k\}^K_{k=1}$ along the $z$-axis of the end-effector, expressed in the object frame, and compute
\begin{equation}
\mathbf{q}_{\text{min}} = \argmin_{\mathbf{q}^k, k \in K} \phi({}^o\mathbf{q}^k),
\end{equation}
\begin{equation}
\Delta\mathbf{q} = \phi({}^o\mathbf{q}_{\text{min}}) \frac{\nabla\phi({}^o\mathbf{q}_{\text{min}})}{\|\nabla\phi({}^o\mathbf{q}_{\text{min}})\|},
\end{equation}
where $\nabla\phi(\mathbf{q})$ denotes the spatial gradient of the SDF.
This yields a translation vector pointing from the object surface toward the sensor axis, with magnitude equal to the signed distance at the closest sampled point. Since we require a planar transform, $\Delta\mathbf{q}$ is projected onto the xy-plane to form $\Delta\mathbf{q}_{xy} = [\Delta q_x, \Delta q_y, 0]^\top$. The translation vector is then adjusted for radius $r_s$ of the cylindrical end-effector 
\begin{equation}
\label{eq:projection}
\Delta\mathbf{q}^*_{xy} = \Delta\mathbf{q}_{xy} - \frac{\Delta\mathbf{q}_{xy}}{\|\Delta\mathbf{q}_{xy}\|}(r_s + \delta),
\end{equation}
with $\delta \sim \mathcal{U}([\delta_{\text{pen}}, \delta_{\text{max}}])$ representing a sampled penetration depth that approximates the compliance of the tactile sleeve. Finally, we apply the resulting translation to project the object into contact with the end-effector, giving the updated configuration $(\tilde{\mathbf{x}}^{o}, \mathbf{u})$ that satisfies~\eqref{eq:contact_constraint}.

\section{Tactile Observation Model}
\label{sec:sensor_model}

We simulate the response of the distributed tactile sensor using a geometric observation model designed to capture essential geometric constraints for object pose estimation. Responses of individual taxels are modeled as functions of signed distance to the object surface, such that combined activations form tactile imprints that can inform downstream estimation through geometric constraints. The full sensor response is obtained by combining individual taxel activations with stochastic effects that model inactive patches and hysteresis. This sensor model is used both for training the inverse model and for likelihood evaluation during filtering.
\begin{figure}[t]
  \centering
  \includegraphics[width=\linewidth]{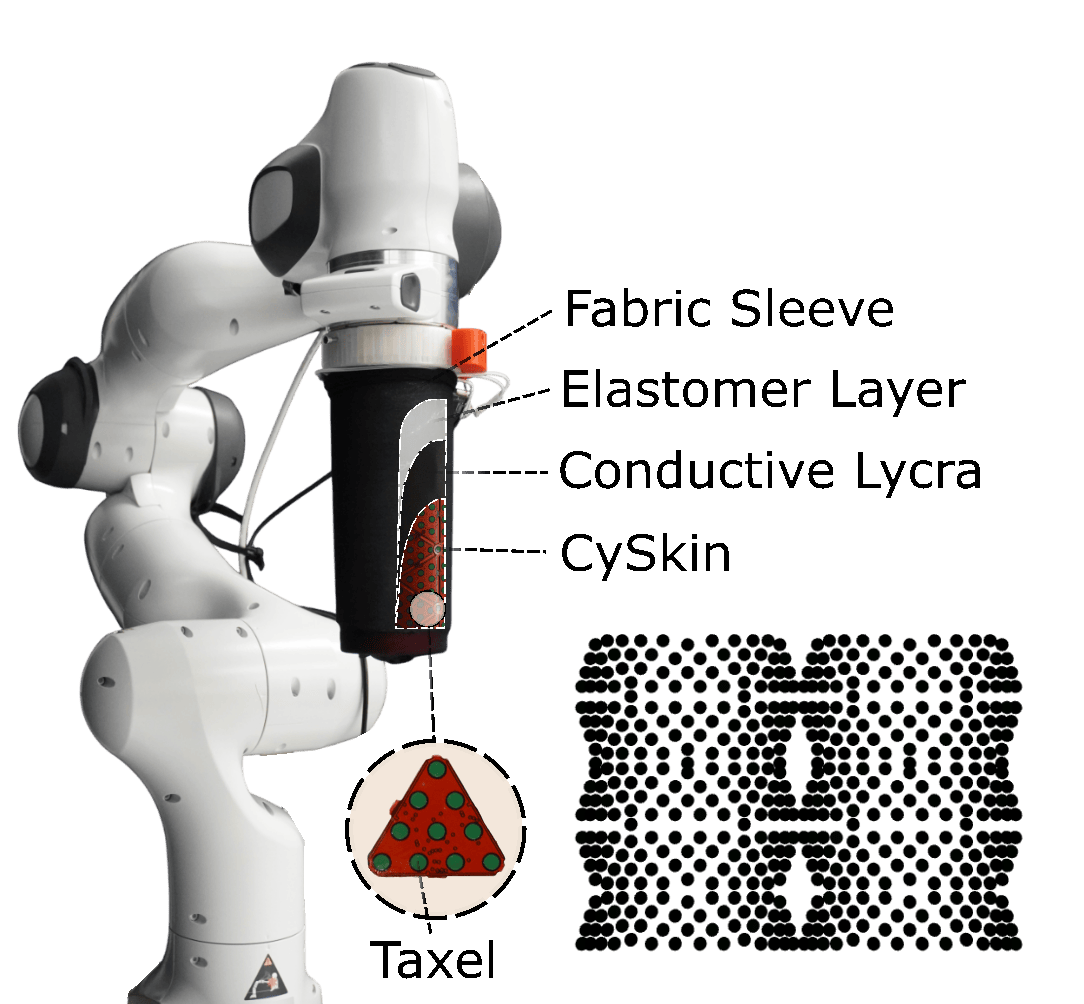}
  \caption{\textit{Left:} cylindrical end-effector mounted on a Franka Emika 7-DoF manipulator and equipped with a distributed tactile sensor (CySkin). The sensor is covered by a conductive fabric layer, followed by a compliant elastomer layer and a fabric sleeve. \textit{Center:} a single triangular skin module. \textit{Right:} full distributed taxel array.}
  \label{fig:setup}
\end{figure}

\begin{figure}[t]
  \centering
  \includegraphics[width=\linewidth]{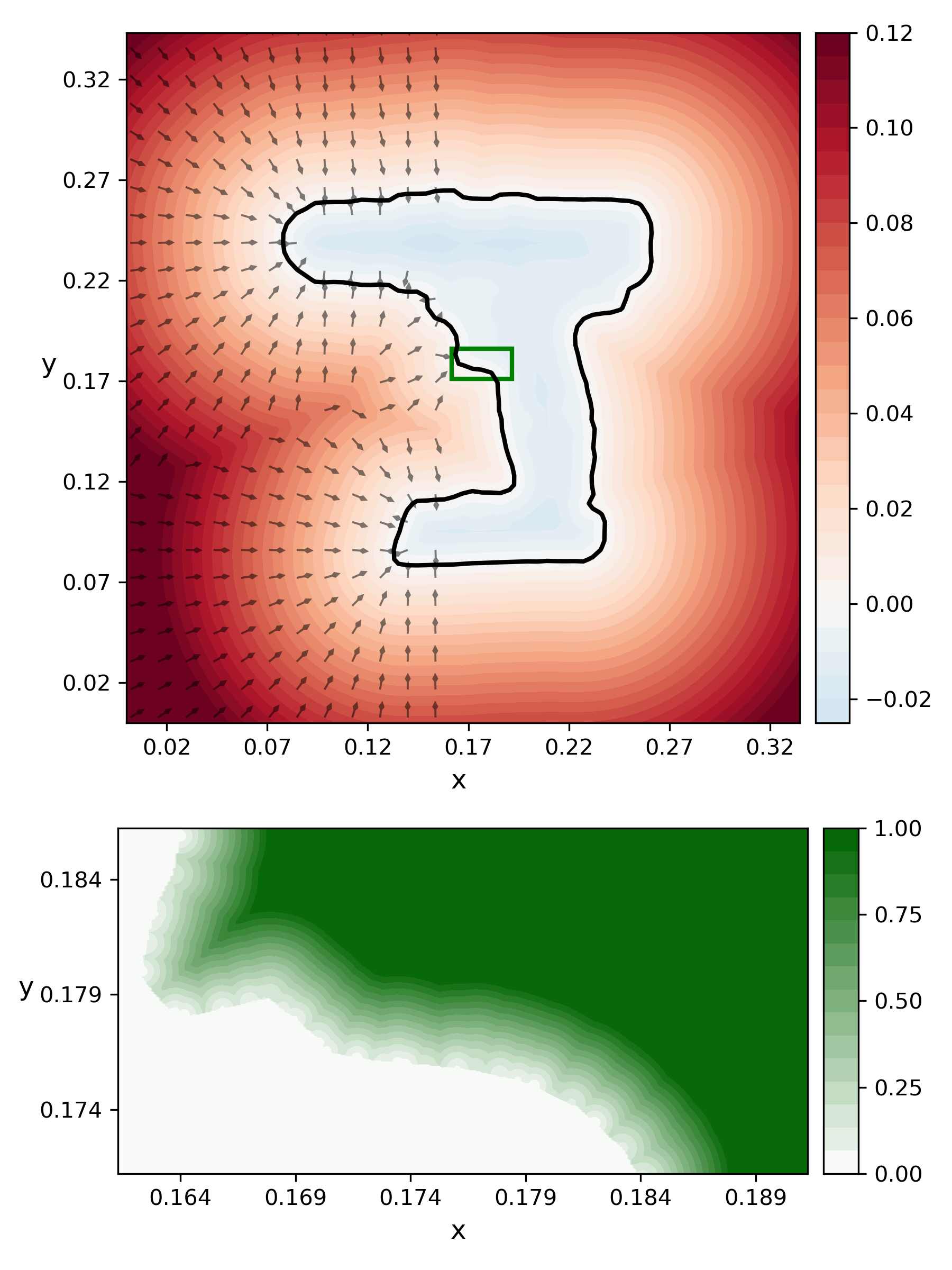}
  \caption{\textit{Top:} 2D slice of the signed distance field (SDF) of a drill object. The colormap shows signed distance values, with the zero-level set (object boundary) indicated by a black contour. Arrows in the left half-plane denote SDF gradients used for contact projection. \textit{Bottom:} Expected taxel activation under the geometric sensor model, shown as a colormap for the region designated by a green box.}
  \label{fig:sensor_model}
\end{figure}
Given an object state $\mathbf{x}^o_t$ and a robot state $\mathbf{u}_t$, the corresponding tactile observation $\mathbf{z}_t$ is modeled as a sample from a conditional distribution of the form
\begin{equation}
\mathbf{z}_t \sim p(\cdot \mid {}^s\mathbf{x}^o_t),
\end{equation}
where ${}^s\mathbf{x}^o_t$ denotes the object pose expressed in the end-effector (sensor) frame, and $\mathbf{z}_t \in \mathbb{R}^{N_{\text{tax}}}$ represents normalized taxel activations.

\subsection{Distance-Based Taxel Model}
For an indexed set of taxels $I = [1, \ldots, N_{\text{tax}}]$, the expected activation of taxel $i \in I$ with center coordinate ${}^o\mathbf{q}^i \in \mathbb{R}^3$ in the object frame, is expressed as a function $f$ of its signed distance to the object surface
\begin{equation}
\mathbb{E}[z^i \mid {}^o\mathbf{q}^i] = f(\phi({}^o\mathbf{q}^i)),
\end{equation}
where $f$ can be any function that captures the distance-based response characteristics of the sensor. We assume that taxels in close proximity to the object surface experience activations proportional to the compression depth of the compliant layers, and define $f$ as a piecewise linear function
\begin{equation}
f(\phi({}^o\mathbf{q}^i)) = \begin{cases}
1 - \phi({}^o\mathbf{q}^i)/d_{\text{max}} & \text{if } \phi({}^o\mathbf{q}^i) < d_{\text{max}}, \\
0 & \text{if } \phi({}^o\mathbf{q}^i) \geq d_{\text{max}},
\end{cases}
\end{equation}
where $d_{\text{max}}$ represents the maximum distance at which activations occur, accounting for the thickness of the soft layer of the sensor. The activation model of a single taxel is visualized in Figure~\ref{fig:sensor_model}.

\subsection{Distributed Array Response Model}
Noisy sensor activations are simulated by sampling from a Gaussian distribution centered at the expected activation $\mu = \mathbb{E}[z^i \mid {}^o\mathbf{q}^i]$ with variance $\sigma^2_{\text{tax}}$. The response $\mathbf{z} = \{z^i\}_{i \in I}$ of the tactile skin is thus obtained by sampling
\begin{equation}
\mathbf{z} \sim \mathcal{N}(\boldsymbol{\mu}, \mathbf{I}\sigma^2_{\text{tax}}),
\end{equation}
for any object and sensor pose.
Taxel responses are assumed conditionally independent given the object and sensor pose. Transient effects such as inactive patches caused by imperfect contact geometry, or hysteresis caused by compression of the tactile sleeve, are modeled by modifying the distribution for a subset $J \subset I$ of taxels
\begin{equation}
\{z^j\}_{j \in J} \sim \mathcal{N}(\tilde{\mu}, \tilde{\sigma}^2_{\text{tax}}),
\end{equation}
where $J$ is selected based on uniformly sampled thresholds on taxel coordinates $\{{}^s\mathbf{q}^j\}_{j \in J}$ in the sensor frame. The biased mean is set to $\tilde{\mu} = 0$ for inactive patches and can be set to $\tilde{\mu} > \mu$ for hysteresis effects, although we handle hysteresis at runtime through signal filtering. The probability of triggering these effects is set empirically based on real-world observations. The resulting forward sensor model allows us to sample
\begin{equation}
    \label{eq:forward_model}
\tilde{\mathbf{z}}_t \sim \tilde{p}(\cdot \mid {}^s\mathbf{x}^o_t)
\end{equation}
to efficiently simulate the response of the tactile array for any object-sensor configuration by leveraging a learned or precomputed SDF.

\section{Diffusion-Based Inverse Sensor Model}
\label{sec:inverse_observation_model}
\begin{figure*}[t]
  \centering
  \includegraphics[width=\linewidth]{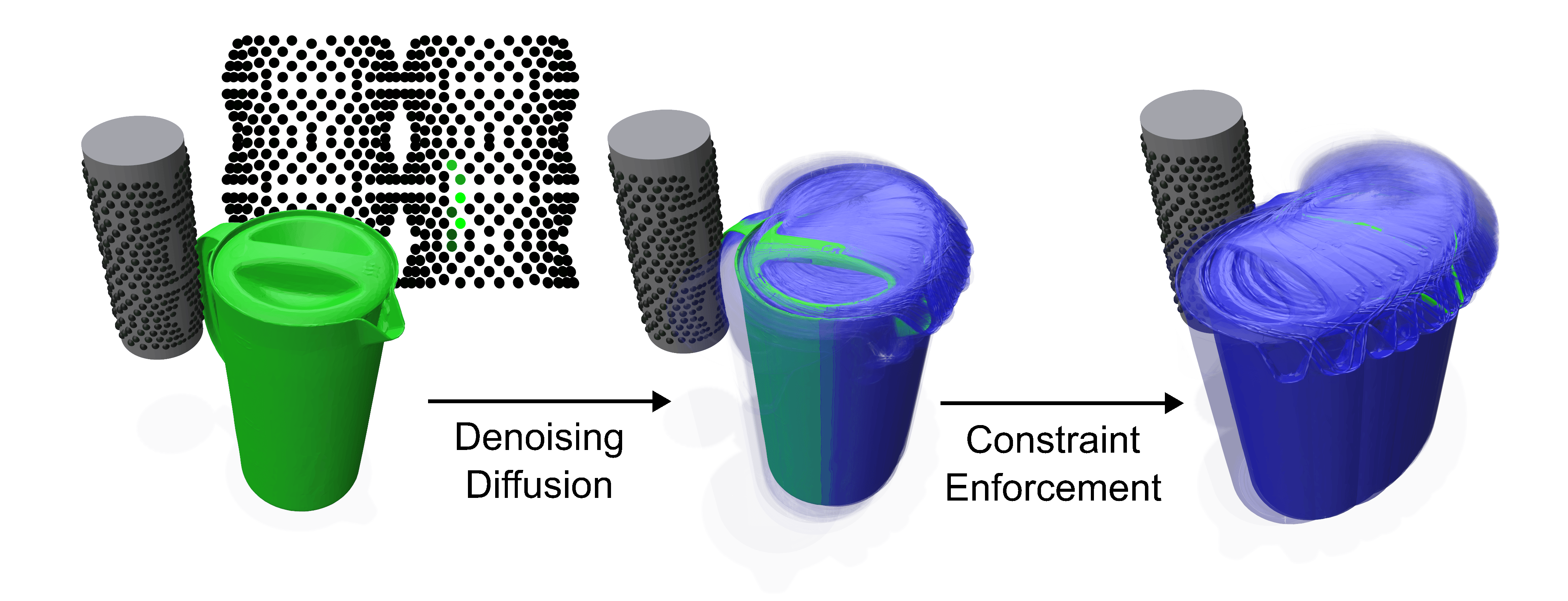}
  \caption{Inference pipeline of the inverse sensor model. \textit{Left:} ground-truth contact configuration and corresponding tactile observation. \textit{Middle:} denoised DDIM particles. \textit{Right:} final pose hypotheses after SDF-based constraint projection. Particles are shown in blue with opacity proportional to likelihood under the observation model. The ground-truth is shown in green.}
  \label{fig:inference_pipeline}
\end{figure*}
To sample proposal particles from the contact manifold based on incoming tactile observations, we learn an inverse model of the distributed tactile sensor. The inverse sensor model $\hat{p}_\theta$ generates a set of multimodal pose hypotheses $\{{}^s\hat{\mathbf{x}}^o_i\}^{N_p}_{i=1}$ for any given tactile observation, where each hypothesis
\begin{equation}
{}^s\hat{\mathbf{x}}^o \sim \hat{p}_\theta({}^s\mathbf{x}^o_t \mid \mathbf{z}_t),
\end{equation}
represents an object pose expressed in the sensor frame. The model combines a diffusion-based generative model with gradient-based projection for contact constraint enforcement during inference.

\subsection{Learning the Inverse Model}

To learn the inverse sensor model, we condition a denoising diffusion model on tactile observations using the contact generation procedure from Section~\ref{sec:contact_generation} and the tactile observation model from Section~\ref{sec:sensor_model}. Given a tactile observation $\mathbf{z}$, the model is trained to produce a set of pose hypotheses expressed in the sensor frame. We rely on a simple denoiser architecture to enable fast training and online sampling in downstream estimators.

\subsubsection{Dataset Generation}

For a known object, we construct a dataset
\begin{equation}
\mathcal{D} = \{({}^s\mathbf{x}^o_n, \mathbf{z}_n)\}^{N_d}_{n=1},
\end{equation}
by uniformly sampling object poses from $\mathcal{X}^o$ and projecting them into contact with the sensor surface using SDF projection~\eqref{eq:projection}. Depending on contact geometry, this can lead to a non-uniform distribution over the contact manifold. Specifically, regions near convex object surfaces tend to be over-represented since more initial poses project to similar contact configurations. To reduce this bias, we apply histogram-based binning over the relative position between the object and the end-effector, and over the object orientation $\theta$, retaining a fixed number of samples per bin. The associated tactile observation $\mathbf{z}_n$ is simulated using the distance-based sensor model~\eqref{eq:forward_model}. 

\subsubsection{Denoiser Architecture and Training}

Following \citet{ho_denoising_2020}, we train the denoiser of a denoising diffusion probabilistic model (DDPM) by progressively adding Gaussian noise to ground-truth poses. The denoiser network receives as input a noisy object pose ${}^s\mathbf{x}^o_{n,t}$, the diffusion timestep $t$, and an array of normalized taxel activations $\mathbf{z} \in [0, 1]^{N_{\text{tax}}}$. It is trained to predict the added noise $\boldsymbol{\epsilon}_{n,t}$ by minimizing the expected mean squared error (MSE) loss
\begin{equation}
\theta = \arg \min_\theta \mathbb{E}_{n,t,\boldsymbol{\epsilon}} \left[ \|\boldsymbol{\lambda}(\boldsymbol{\epsilon}_\theta({}^s\mathbf{x}^o_{n,t}, \mathbf{z}_n, t) - \boldsymbol{\epsilon}_{n,t})\|^2 \right],
\end{equation}
where $\boldsymbol{\lambda}$ is a diagonal matrix that balances translational and rotational state components, and $\lVert \cdot \rVert$ denotes the Euclidean norm.

\subsection{Geometry-Aware Inference}
At inference time, we generate pose hypotheses by applying denoising diffusion implicit model (DDIM) sampling~\citep{song_denoising_2022} using the trained noise predictor $\boldsymbol{\epsilon}_\theta$, followed by geometric constraint enforcement. DDIM is consistent with the DDPM forward process used during training but allows sampling over a reduced set of timesteps, enabling faster inference with minimal loss in sample quality.

Given a tactile observation $\mathbf{z}$, an initial latent pose ${}^s\hat{\mathbf{x}}^o_{\tau_S} \sim \mathcal{N}(\mathbf{0}, \mathbf{I})$ is iteratively denoised over $S$ timesteps. At each timestep $\tau_i$, the denoiser predicts the added noise $\hat{\boldsymbol{\epsilon}} = \boldsymbol{\epsilon}_\theta({}^s\hat{\mathbf{x}}^o_{\tau_i}, \mathbf{z}, \tau_i)$ and updates the pose hypothesis. Running this procedure for $N_p$ initial noise samples produces a set $\{{}^s\hat{\mathbf{x}}^o_i\}^{N_p}_{i=1}$ of pose hypotheses consistent with the observation $\mathbf{z}$. Additional details on the utilized model architecture, training, and inference procedures are provided in Appendix~\ref{app:diffusion}.

\subsubsection{Constraint Enforcement via SDF Projection}

Although the learned inverse sensor model produces poses that lie close to the contact manifold, small constraint violations can cause significant errors in downstream estimation due to the highly nonlinear relationship between object pose and tactile observations. We therefore ensure that all generated hypotheses satisfy contact constraints~\eqref{eq:contact_constraint} by applying an SDF projection step~\eqref{eq:projection} to each sample. This produces the final set $\{{}^s\hat{\mathbf{x}}^o_i\}^{N_p}_{i=1}$ of candidate poses that satisfy contact constraints, preventing downstream estimation errors.
Explicitly enforcing contact constraints also improves sample efficiency, as all propagated particles remain valid hypotheses. In practice, the computational overhead is negligible since SDF values and gradients can be precomputed or modeled using function approximators. The bias introduced by projection is also minimal, as the required corrections are typically small. The inference pipeline is illustrated in Figure~\ref{fig:inference_pipeline}.

\section{Particle Filter with Belief-Informed Injection}
\label{sec:particle_filter}

To estimate object pose over sequential contacts, we integrate the learned inverse sensor model with a particle filter, which maintains a non-parametric belief that is updated over multiple contacts using incoming observations and robot poses. The belief at time $t$ is represented as a set of $N$ weighted particles
\begin{equation}
\mathcal{X}_{t} = \{\langle \mathbf{x}^{o,[n]}_{t}, w^{[n]}_{t} \rangle\}^N_{n=1},
\end{equation}
approximating the posterior $p(\mathbf{x}^o_{t} \mid \mathbf{z}_{1:t}, \mathbf{u}_{1:t})$ over object pose given all previous observations and actions. As illustrated in Figure~\ref{fig:framework}, the inverse model generates pose hypotheses conditioned on tactile observations, which are injected into the particle filter using a belief-informed proposal scheme. 

\subsection{Bayesian Estimation Loop}

The belief is updated using a standard Bayesian filtering procedure consisting of prediction, measurement update, and resampling steps.

\paragraph{Prediction.}
During each contact, particle poses are propagated using the transition model
\begin{equation}
\mathbf{x}^{o,[n]}_t \sim p(\mathbf{x}^o_t \mid \mathbf{x}^{o,[n]}_{t-1}, \mathbf{u}_t),
\end{equation}
producing a set $\bar{\mathcal{X}}_t$ of predicted pose particles for timestep $t$. In static settings, the transition model simplifies to a Dirac delta $\delta(\mathbf{x}^o_t - \mathbf{x}^o_{t-1})$, retaining previous poses exactly.

\paragraph{Measurement update.}
The observation likelihood of each predicted pose particle is computed as a product over all taxels using the distance-based sensor model~\eqref{eq:forward_model}
\begin{equation}
\mathcal{L}_{\text{obs}}(\mathbf{z}_t \mid {}^s\mathbf{x}^{o,[n]}_t)
=
\prod_{i=1}^{N_{\text{tax}}}
\tilde{p}(z^i \mid {}^s\mathbf{x}^{o,[n]}_t).
\end{equation}
Likelihoods are calculated for each particle to update the weights of the predicted belief $\bar{\mathcal{X}}_t$ based on the current tactile observation
\begin{equation}
\tilde w^{[n]}_t
=
w^{[n]}_{t-1}\,
\mathcal{L}_{\text{obs}}(\mathbf{z}_t \mid {}^s\mathbf{x}^{o,[n]}_t).
\end{equation}
To mitigate particle depletion, we use a modified likelihood in the particle filter, with variance dependent on taxel distance from the object surface:
\begin{equation}
\tilde{\sigma} = \tilde{\sigma}_{\min} + \frac{\tilde{\sigma}_{\max} - \tilde{\sigma}_{\min}}{1 + \exp\bigl(\kappa(\phi({}^o\mathbf{q}^i) - d_0)\bigr)},
\end{equation}
where $^o\mathbf{q}^i$ represents the position of taxel $i$ in the object frame, $\kappa$ controls the transition sharpness and $d_0$ is an offset from the surface. This assigns higher variance to taxels in contact, where model inaccuracies are more pronounced.

\paragraph{Resampling.}
Following the measurement update, weights are normalized to satisfy $\sum_{n=1}^N w^{[n]} = 1$, and low-variance resampling (LVR) \citep{thrun_probabilistic_nodate} is applied when the effective sample size
\begin{equation}
\text{ESS} = \frac{1}{\sum_{n=1}^N (w^{[n]})^2}
\end{equation}
falls below a threshold, with $\text{ESS} \in [1, N]$ for normalized weights.
This discards particles that are inconsistent with recent observations.

\begin{algorithm}[t]
\caption{Belief-Informed Particle Injection}
\label{alg:pf}
\KwIn{Predicted belief $\bar{\mathcal{X}}_{t}$, observation $\mathbf{z}_t$, sensor pose $\mathbf{u}_t$}
\KwOut{Updated belief $\mathcal{X}_t$}

\ForEach{particle $\langle \bar{\mathbf{x}}^o, \bar{w}_{t} \rangle \in \bar{\mathcal{X}}_t$}{
$\bar{w}_t \leftarrow \bar{w}_{t}\,
\mathcal{L}_{\text{obs}}(\mathbf{z}_t \mid {}^s\bar{\mathbf{x}}^{o}_t)
$\;
}

$\hat{\mathcal{X}}_t \leftarrow \{\langle \hat{\mathbf{x}}^{o,[n]}, \hat{w}^{[n]}_t \rangle\}_{n=1}^{N_p},
\quad
{}^s\hat{\mathbf{x}}^o \sim \hat{p}_\theta({}^s\mathbf{x}^o \mid \mathbf{z}_t)$\;

\ForEach{hypothesis $\langle \hat{\mathbf{x}}^o, \hat{w}_t \rangle \in \hat{\mathcal{X}}_t$}{
$\ell_t \leftarrow \log\mathcal{L}_{\mathrm{obs}}(\mathbf{z}_t \mid {}^s\hat{\mathbf{x}}^o)
+ \tilde{\ell}_t(\hat{\mathbf{x}}^o)$\;
$\hat{w}_t \leftarrow \exp(\ell_t)$\;
}

$\mathcal{Y}_t \leftarrow \bar{\mathcal{X}}_t \cup \hat{\mathcal{X}}_t$\;
$\mathcal{X}_t \leftarrow \textsc{LVR}(\mathcal{Y}_t)$\;
\Return{$\mathcal{X}_t$}
\end{algorithm}

\subsection{Belief-Informed Particle Injection}
During each contact, new tactile observations are integrated by injecting pose hypotheses generated by the learned inverse sensor model into the existing belief.
To produce a weighted set of candidate particles
$\hat{\mathcal{X}}_t=\{\langle \hat{\mathbf{x}}^{o,[n]}_t, \hat{w}^{[n]}_{t}\rangle\}_{n=1}^{N_p}$,
we first sample object pose hypotheses from the inverse model
\begin{align}
    {}^s\hat{\mathbf{x}}^o_t
    \sim
    \hat{p}_\theta({}^s\mathbf{x}^o_t \mid \mathbf{z}_t),
\end{align}
and transform them to the global frame based on the known end-effector pose $\mathbf{u}_t$. As the inverse sensor model conditions hypotheses only on the current observation, we incorporate information from the existing belief during proposal through an additional consistency term that encourages hypotheses to lie near regions supported by previous observations.

Each candidate pose \(\hat{\mathbf{x}}^o_t\) is assigned a consistency score based on its proximity to the \(k\) nearest neighbors in the previous pose set \(\bar{\mathcal{X}}_t\)
\begin{align}
\label{eq:consistency_term}
\tilde{\ell}_t(\hat{\mathbf{x}}^o_t)
=
\frac{
    \sum_{i \in \mathcal{N}_k(\hat{\mathbf{x}}^o_t)}
    w^{[i]}_{t-1}\,
    \mathcal{K}_h\!\left(\Delta \mathbf{x}_i\right)
}{
    \sum_{i \in \mathcal{N}_k(\hat{\mathbf{x}}^o_t)} w^{[i]}_{t-1}
},
\end{align}
where \(\mathcal{N}_k(\hat{\mathbf{x}}^o_t)\) denotes the index set of the \(k\) nearest neighbors of \(\hat{\mathbf{x}}^o_t\) in \(\bar{\mathcal{X}}_t\).
The similarity between poses is measured using a Gaussian kernel and pose difference vector \(\Delta \mathbf{x}\), with angular differences computed on the unit circle,
\begin{align}
\label{eq:kernel_function}
\mathcal{K}_h\left(\Delta\mathbf{x}\right)
=
\exp\!\left(
    -\tfrac{1}{2}\|\boldsymbol{\lambda}\,\Delta \mathbf{x}\|^2 / h^2
\right),
\end{align}
where $\lVert\cdot\rVert$ denotes a Euclidean norm and $\boldsymbol{\lambda}$ balances translational and rotational components.
This term provides a non-parametric estimate of local support under the current belief and discourages injected samples that fall in regions inconsistent with prior observations. To prevent oversmoothing and preserve multimodal structure, a small value of \(k\) is used. As contacts accumulate, emphasis is gradually shifted from exploration to consistency with prior beliefs by exponentially reducing the kernel bandwidth $h$ within predefined bounds.

Each injected particle is assigned an unnormalized log-score by combining the observation log-likelihood with the consistency term,
\begin{align}
\ell_{t}
=
\log\mathcal{L}_{\mathrm{obs}}(\mathbf{z}_t\mid {}^s\hat{\mathbf{x}}^o_t)
+
\tilde{\ell}_t(\hat{\mathbf{x}}^o_t),
\end{align}
and the final weight obtained as $\hat{w}_t = \exp(\ell_{t})$.
The weighted set $\hat{\mathcal{X}}_t$ is then combined with the predicted particle set $\bar{\mathcal{X}}_t$ to form a unified set of proposal particles
\begin{align}
    \mathcal{Y}_t
    =
    \bar{\mathcal{X}}_t \cup \hat{\mathcal{X}}_t,
\end{align}
which preserves high-probability modes from the prediction step while introducing new hypotheses conditioned on the current tactile observation.
Low-variance resampling is finally applied over $\mathcal{Y}_t$, using the normalized scores, to produce an updated particle set of size $N$ that approximates
$p(\mathbf{x}^o_t \mid \mathbf{z}_{1:t},\mathbf{u}_{1:t})$.
The particle injection procedure is summarized in Algorithm~\ref{alg:pf}.

\section{Experiments}
\label{sec:experiments}
We design our experiments to evaluate the proposed framework at three levels: (i) the inverse tactile sensor model and geometry-aware contact inference in isolation (Section~\ref{subsec:sample_efficiency}), (ii) their integration within a particle filtering framework for pose estimation (Section~\ref{subsec:static_object}), and (iii) robustness under dynamic contact interactions and real-world sensing conditions (Section~\ref{subsec:box_pushing}). We evaluate displacement error using ADD and ADD-S metrics on a test set of household objects from the YCB dataset~\citep{YCBds} with varying levels of tactile discriminability, ranging from objects with distinctive geometric features (power drill) to those with flat, featureless surfaces (bulky box). For each experiment type, we present both simulated and real-world results to evaluate robustness across domains.

\subsection{Experimental Setup}
\label{subsec:exp_setup}
\begin{figure}[t]
  \centering
  \includegraphics[width=\linewidth]{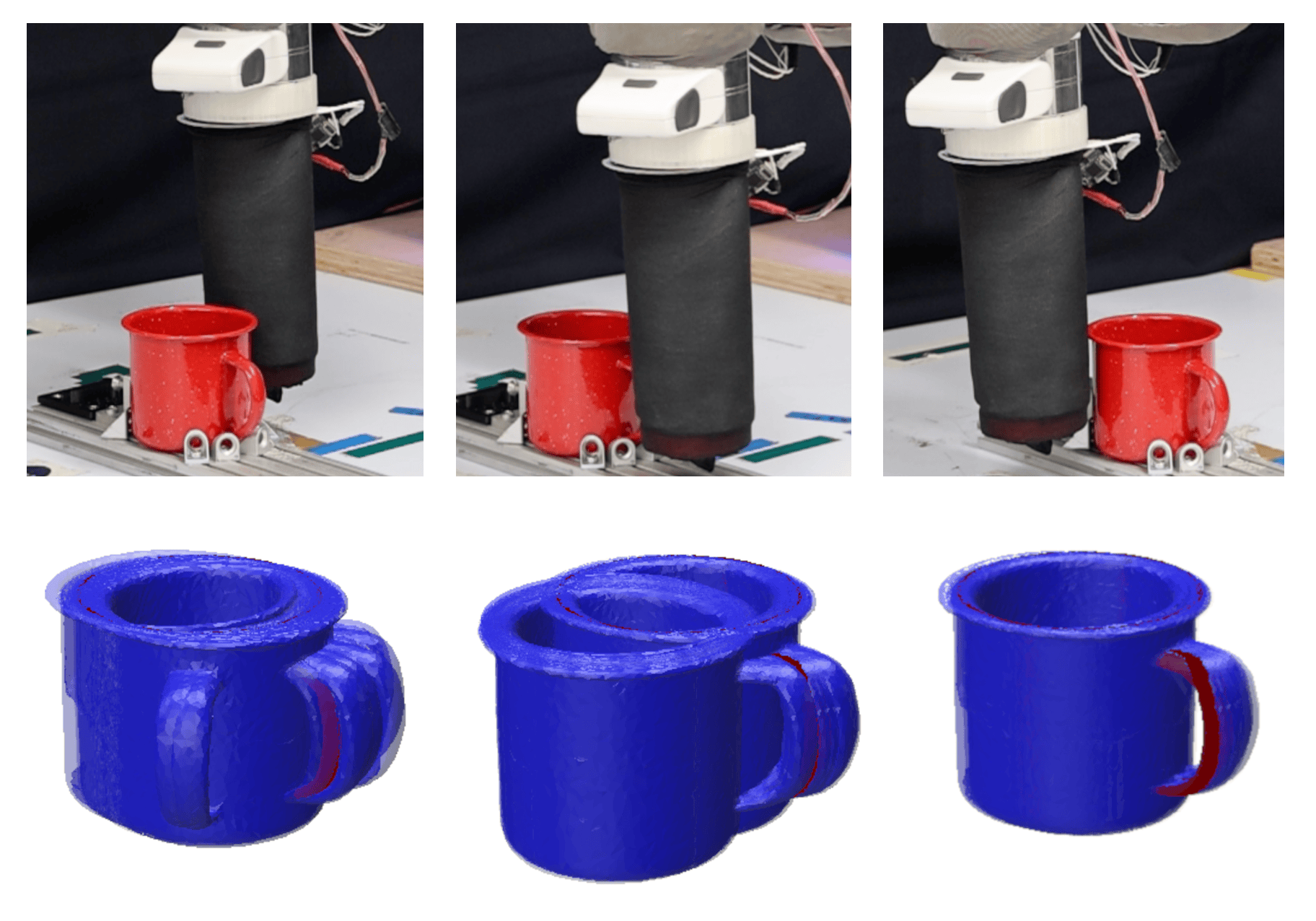}
  \caption{Belief evolution during real-world pose estimation. Over a sequence of contacts with a static mug, the particle filter integrates hypotheses from the inverse sensor model trained in simulation. Belief particles shown in blue, with opacity proportional to weight.}
  \label{fig:mug_real}
\end{figure}

\subsubsection{Hardware Platform and Sensor Configuration}
Our setup consists of a 7-DoF Franka Emika manipulator equipped with a cylindrical end-effector covered by a distributed tactile sensor based on CySkin technology~\citep{maiolino_flexible_2013}. The sensor consists of a flexible PCB (fPCB) with capacitive taxels, covered by conductive Lycra as the second conductive layer. An elastomer layer enables conforming to local object geometry upon contact and acts as a spatial filter, with a fabric sleeve preventing direct contact between the elastomer and objects. The complete setup is shown in Figure~\ref{fig:setup}. The sensor has a non-uniform taxel distribution with a density of 1.56 taxels/$cm^2$ and 514 total taxels. For resolution studies specifically, taxels are uniformly instantiated at preset densities. At runtime, tactile observations are provided at 20\,Hz and normalized to $[0, 1]$ based on calibration values, with activations below a threshold set to zero to filter noise.

\subsubsection{Simulation Environment and Implementation}
We simulate contact using SDF projection (Section~\ref{sec:contact_generation}) with the tactile observation model (Section~\ref{sec:sensor_model}). The end-effector is randomly rotated around the $z$-axis, which is perpendicular to the ground plane. Simulated observations are drawn from~\eqref{eq:forward_model}, and domain randomization applied through simulated inactive patches to model imperfect contact conditions. Pushing experiments are implemented in MuJoCo~\citep{todorov_mujoco_2012}, providing contact dynamics not explicitly modeled during training.

\subsubsection{Test Objects and Model Training}
Experiments use household objects from the YCB dataset with varying geometric complexity. Numerical analyses consider three representative objects: the power drill (distinctive features), mug (moderate complexity), and mustard bottle (simple geometry). For additional variety, we include a featureless box mesh and a real-world drill reconstructed from scanned point clouds. The full list of objects is provided in Appendix~\ref{app:objects}. Inverse sensor models are trained per-object using the procedure from Section~\ref{sec:inverse_observation_model}, with uniform coverage in the training set achieved via histogram-based binning over contact configurations. 

\subsection{Baselines and Metrics}

\subsubsection{Baseline Methods}
We compare against local sampling baselines that explore the space of hypotheses through perturbations around existing particles. After applying the measurement update, the baseline resamples the non-parametric belief using low-variance resampling, then applies uniformly sampled noise to explore hypotheses around high-likelihood particles. Perturbation bounds are progressively tightened as contacts accumulate, and SDF projection enforces contact constraints in all generated hypotheses. Additionally, two variants of measurement updates are evaluated: (1) full tactile skin for measurement updates, and (2) simulated force-torque sensor using averaged responses from active taxels. Baseline algorithm details are provided in Appendix~\ref{app:baseline}.

\subsubsection{Evaluation Metrics}
We use average distance of model points (ADD) and ADD-S for symmetric objects to evaluate pose estimation accuracy
\begin{equation}
\text{ADD} = \frac{1}{|\mathcal{M}|} \sum_{\mathbf{q} \in \mathcal{M}} \|(\mathbf{R}\mathbf{q} + \mathbf{t}) - (\mathbf{R}^*\mathbf{q} + \mathbf{t}^*)\|,
\end{equation}
where $\mathcal{M}$ denotes points sampled from the object mesh, and $(\mathbf{R}, \mathbf{t})$ and $(\mathbf{R}^*, \mathbf{t}^*)$ are the estimated and ground-truth transforms, respectively. All distances are computed using the Euclidean norm. For objects with symmetries, ADD-S compares each transformed point to the closest point in the ground-truth set
\begin{equation}
\text{ADD-S} = \frac{1}{|\mathcal{M}|} \sum_{\mathbf{q} \in \mathcal{M}} \min_{\mathbf{q}^* \in \mathcal{M}^*} \|(\mathbf{R}\mathbf{q} + \mathbf{t}) - \mathbf{q}^*\|,
\end{equation}
where $\mathcal{M}^* = \{\mathbf{R}^*\mathbf{q} + \mathbf{t}^* \mid \mathbf{q} \in \mathcal{M}\}$ represents the transformed ground-truth points. All reported ADD values are normalized by object diameter $d_{\text{obj}}$ to enable comparison across objects
\begin{equation}
\text{ADD}_{\text{norm}} = \frac{\text{ADD}}{d_{\text{obj}}}.
\end{equation}
Values below 0.1 are considered to accurately estimate the ground truth pose, and are used as a measure of successful convergence. Due to non-normal error distributions arising from bimodal outcomes (successful convergence vs. failure), results are reported as violin plots or numerically using median and interquartile range (IQR). Convergence reliability is quantified using success rates, reporting the number of runs that achieve ADD below the threshold.

\subsection{Sample Efficiency Analysis}
\label{subsec:sample_efficiency}
We first evaluate the inverse sensor model and geometry-aware inference in isolation, without temporal belief propagation. This experiment assesses sample efficiency and coverage of valid pose hypotheses.

Sampling efficiency is evaluated using pose hypotheses generated during initial contact between the sensorized end-effector and objects from the YCB dataset. For each object, we compute the average log-likelihood of ground-truth observations
\begin{equation}
\bar{\mathcal{L}}_{\text{obs}} = \frac{1}{N_pN_g} \sum_{i=1}^{N_g} \sum_{j=1}^{N_p} \mathcal{L}_{\text{obs}}(\mathbf{z}_g \mid {}^s\mathbf{x}^o_{i, j}),
\end{equation}
where $N_p$ is the number of pose hypotheses generated for each of $N_g$ ground-truth observations. Higher average log-likelihoods indicate samples more consistent with observed tactile signals. To study the effect of observation discriminability on sample efficiency, coverage is evaluated by the ADD of maximum-a-posteriori (MAP) hypotheses across contact configurations for multiple objects and sensor resolutions.

\subsubsection{Results}
\begin{figure}[t]
  \centering
  \includegraphics[width=\linewidth]{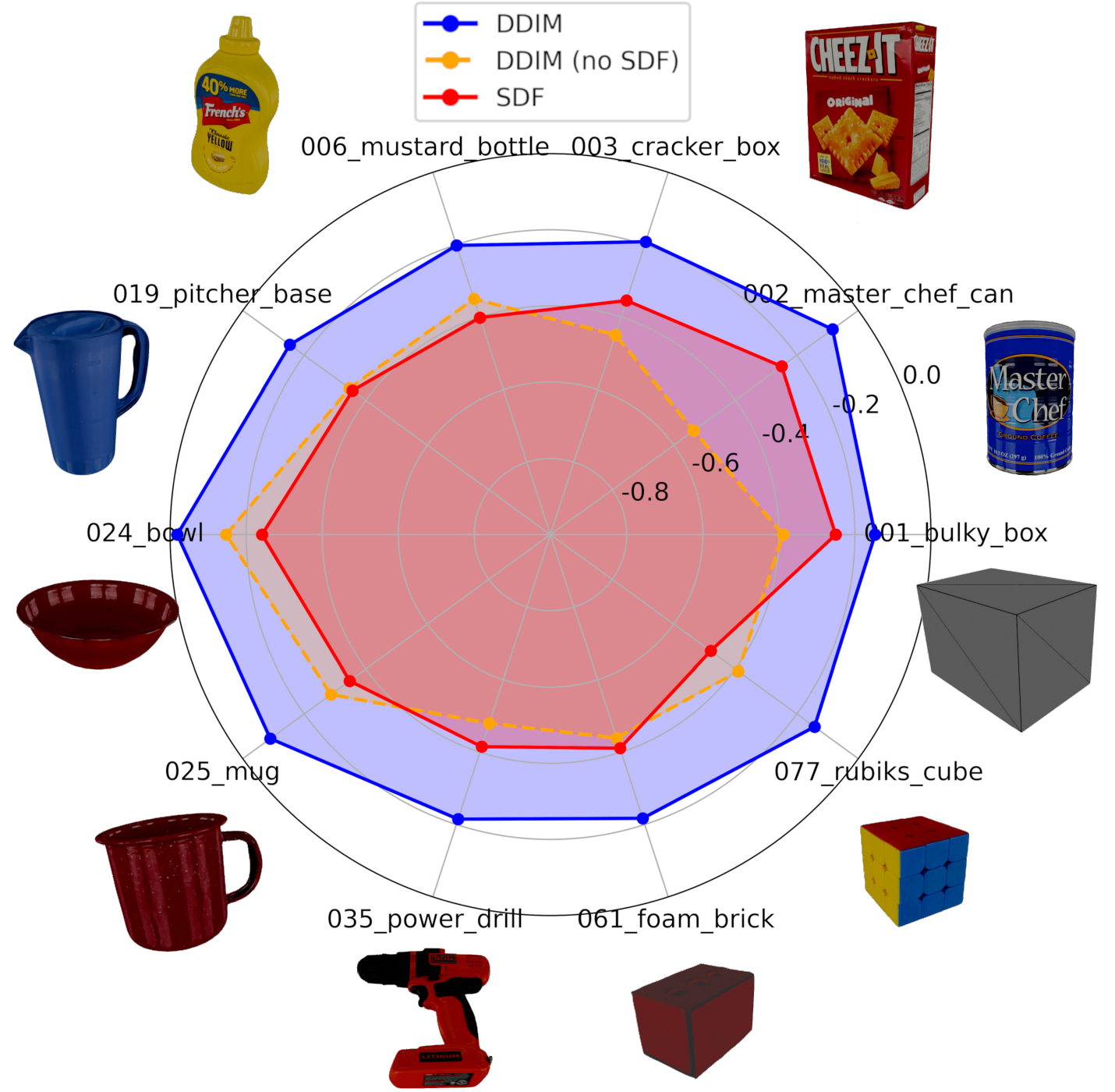}
  \caption{Average log-likelihood of generated samples across test objects. For each object, 100 samples are generated for 100 ground-truth contact configurations. Results compare the learned inverse model with and without SDF constraint enforcement against the local sampling baseline. Higher values indicate greater consistency with ground-truth observations.}
  \label{fig:radar_chart}
\end{figure}
Average log-likelihoods for each object are shown in Figure~\ref{fig:radar_chart}. The learned inverse sensor model (DDIM) produces samples with higher observation consistency than the local sampling baseline (SDF) across all objects. Notably, removing SDF-guided constraint enforcement from the inverse sensor model substantially degrades sample quality, as physically inconsistent hypotheses are penalized heavily by the observation likelihood model.
\begin{figure}[t]
  \centering
  \includegraphics[width=0.9\linewidth]{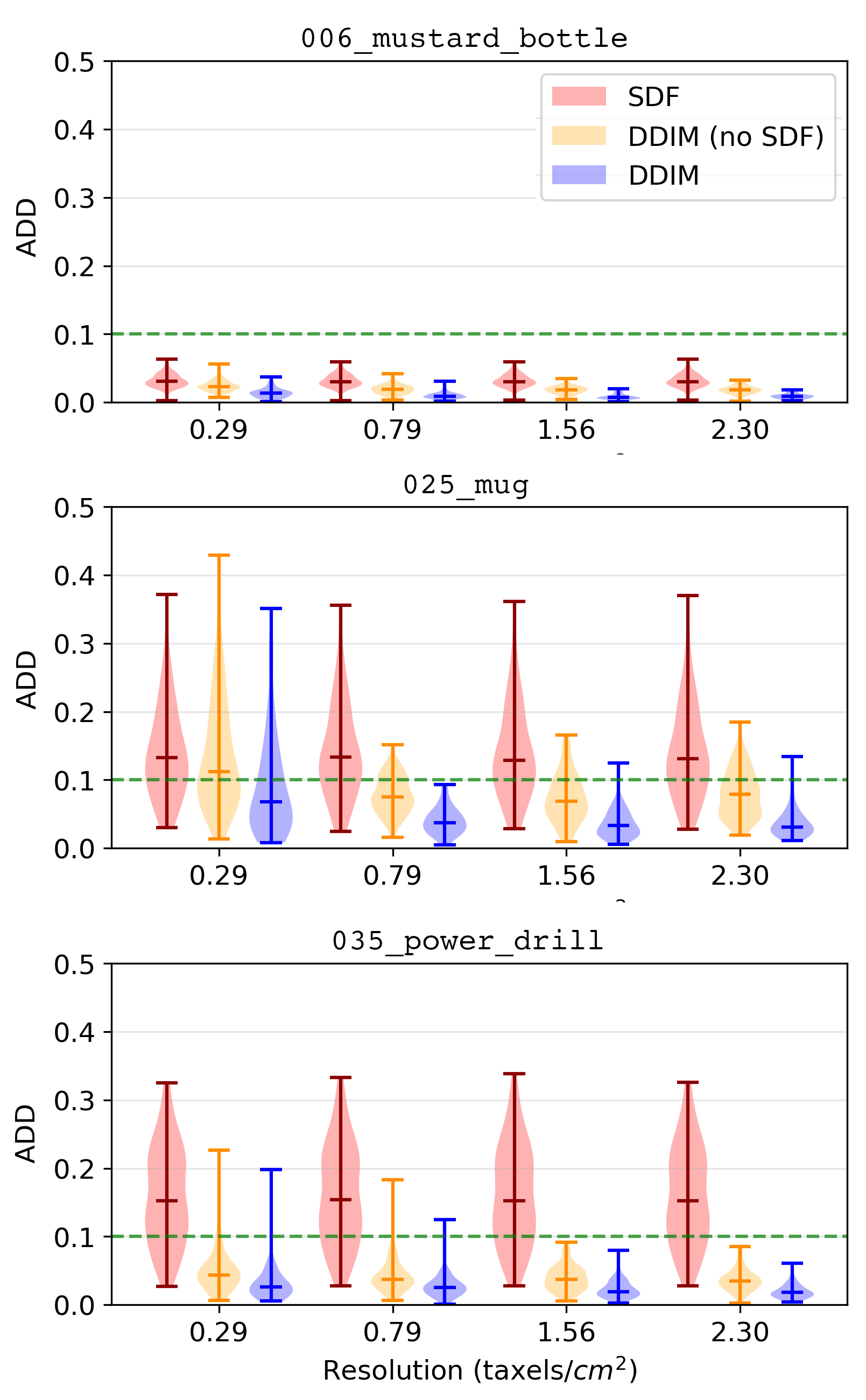}
  \caption{ADD of MAP hypotheses across sensor resolutions. For each contact configuration, the MAP hypothesis is selected from 100 generated samples. Results compare the learned inverse model with and without SDF projection against the local sampling baseline. Lower values indicate higher accuracy, with the 0.1 accuracy threshold shown as a dashed line.}
  \label{fig:taxel_density}
\end{figure}
Figure~\ref{fig:taxel_density} shows ADD distributions of MAP hypotheses for 100 contact configurations, where MAP hypotheses are selected from 100 samples per ground-truth contact. Results are shown for three objects with varying geometric complexity and different taxel densities. Both with and without SDF guidance, the learned inverse model generates hypotheses nearer to ground-truth poses when compared to the local sampling baseline. This advantage is consistent across all tested objects and sensor resolutions. Higher resolutions provide more discriminative observations, though the benefit diminishes beyond 1.56 taxels/$cm^2$ for the tested geometries. For objects with distinctive geometry, such as the power drill, the improved sample efficiency enables accurate hypothesis generation from a single contact, which is not achieved by local sampling.
\begin{table}[h]
\centering
\footnotesize
\begin{tabular}{|c|c|c|c|c|}
\hline
res/$N_p$ & 1 & 10 & 100 & 1000 \\ \hline
0.29 & $18.7 \pm 1.7$ & $21.2 \pm 2.0$ & $30.9 \pm 1.06$ & $74.1 \pm 13.9$ \\ \hline
0.79 & $18.9 \pm 1.4$ & $22.9 \pm 1.6$ & $30.0 \pm 11.2$ & $84.7 \pm 14.3$ \\ \hline
1.56 & $19.8 \pm 2.3$ & $27.5\pm5.1$  & $42.8\pm12.4$ & $134.7\pm24.2$ \\ \hline
2.30 & $20.1\pm2.2$  & $30.0 \pm 4.8$ & $45.3\pm14.6$ & $161.3\pm 27.9$ \\ \hline
\end{tabular}
\caption{Average inference time (ms) for varying sample counts and sensor resolutions. Measured on a 2.80\,GHz Intel Core i7-1165G7.}
\label{tab:inference}
\end{table}
Sampling times for different batch sizes and resolutions are shown in Table~\ref{tab:inference}, demonstrating real-time hypothesis generation at 20\,Hz on a consumer-grade CPU. Computation time increases with sample size and sensor resolution. Additional numerical results and ablation studies are provided in Appendix~\ref{app:results}.

\subsection{Static Object Pose Estimation}
\label{subsec:static_object}
This experiment evaluates the effectiveness of belief-informed particle injection for system-level Bayesian pose estimation in both simulation and real-world conditions. We integrate the learned inverse sensor model in a particle filter using the particle-injection scheme described in Section \ref{sec:particle_filter}. We measure the displacement error (ADD) of the evolving belief over multiple contacts with a static object, starting from a uniform pose prior over the entire planar workspace.

\subsubsection{Simulated Results}
For each object, we run multiple estimation sequences consisting of random contact configurations and corresponding tactile observations. An additional ablation baseline, denoted as SDF (FT), evaluates simulated force-torque responses by averaging active taxel readings to distinguish the benefits of observational discriminability from proposal quality.
\begin{figure}[t]
  \centering
  \includegraphics[width=0.9\linewidth]{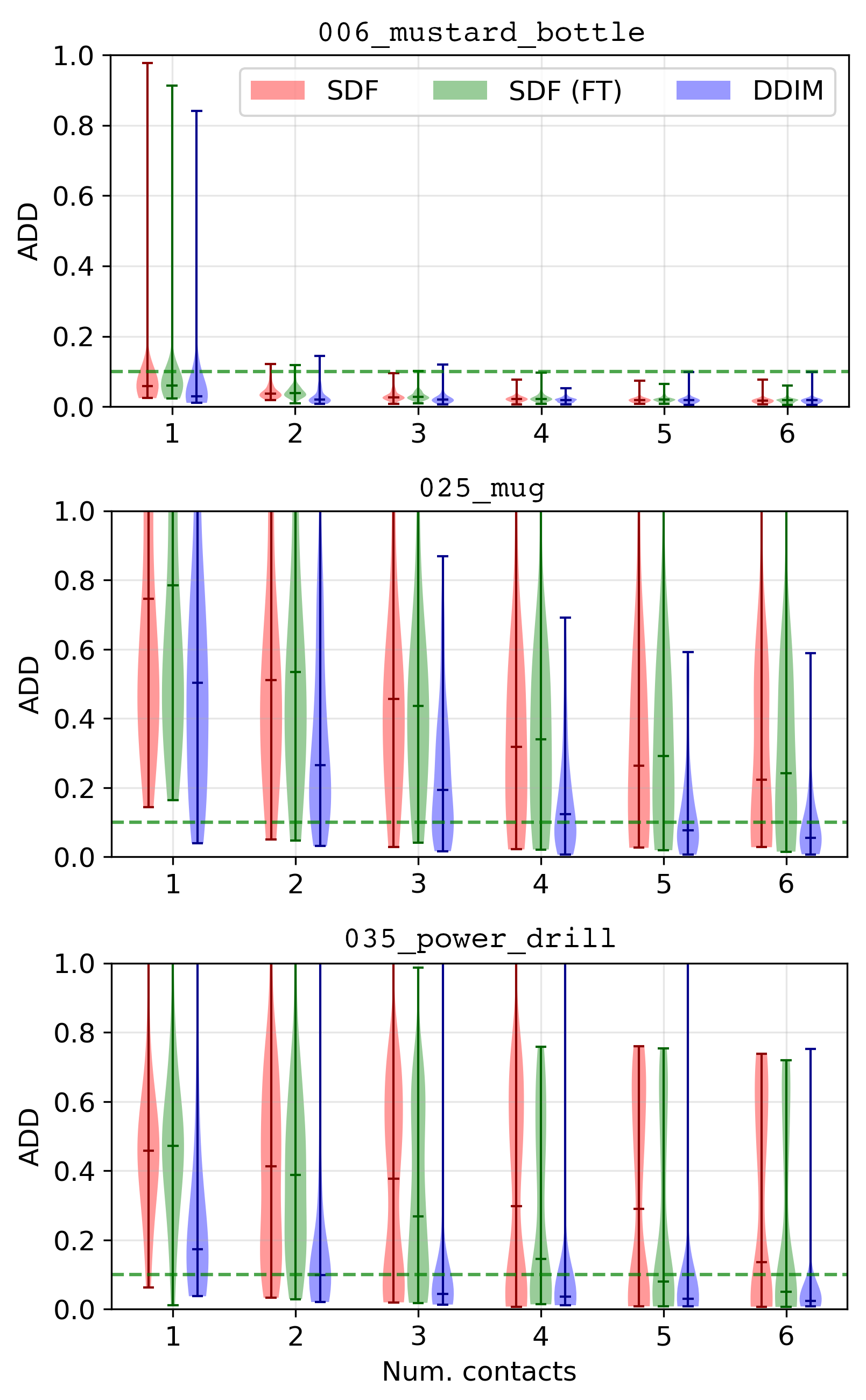}
  \caption{Static pose estimation in simulation. ADD of the averaged particle filter belief is shown over sequences of 6 random contacts for 100 episodes. The learned inverse model is compared against local sampling baselines across objects with varying geometric complexity. Lower values indicate higher accuracy, with the 0.1 threshold indicated by a dashed line.}
  \label{fig:planar_estim}
\end{figure}

Distributions of ADD for the averaged belief over 100 sequences of 6 random contacts are shown in Figure~\ref{fig:planar_estim}. The learned inverse model enables more reliable convergence, with most runs crossing the accuracy threshold within fewer contacts than local sampling baselines. For objects with complex geometries (mug, power drill), local sampling baselines exhibit degraded convergence and multimodal ADD distributions, indicating convergence to incorrect estimates and insufficient coverage of the hypothesis space.

\subsubsection{Real-World Validation}

We conduct pose estimation experiments on the same objects using the teleoperated setup introduced in Section~\ref{subsec:exp_setup}. Models are trained purely on simulated data with domain randomization through inactive patch simulation. Tactile data is collected in 5 runs per object with contacts distributed around the surface, and estimation is run 10 times per recording to account for estimator variance.
\begin{table}[t]
    \centering
    \caption{Static pose estimation in real-world experiments. ADD ($\times 10^{-2}$) of the averaged belief reported as median and IQR for varying numbers of contacts. Values below 10 indicate successful convergence to ADD $< 0.1$. Success rates are shown for 50 runs per object.}
    \label{tab:rw_static}
    \begin{tabular}{|c|c|c|c|}
        \hline
        \multicolumn{4}{|c|}{\texttt{006\_mustard\_bottle}} \\ \hline
        n & DDIM & SDF & SDF (FT) \\ \hline
        1 & 8.30 (1.61) & \textbf{7.06 (1.76)} & 7.15 (1.95) \\ \hline
        3 & \textbf{3.59 (2.06)} & 4.14 (2.08) & 4.51 (2.94) \\ \hline
        5 & 3.97 (1.15) & \textbf{3.83 (1.58)} & 4.60 (3.22) \\ \hline
        7 & \textbf{3.80 (2.00)} & 3.92 (1.92) & 4.43 (3.20) \\ \hline
        Succ. & \textbf{50/50} & 48/50 & 47/50 \\ \hline
        \hline
        \multicolumn{4}{|c|}{\texttt{025\_mug}} \\ \hline
        n & DDIM & SDF & SDF (FT) \\ \hline
        1 & 40.54 (15.55) & 38.56 (12.33) & \textbf{36.73 (11.32)} \\ \hline
        3 & \textbf{17.76 (13.33)} & 32.91 (40.20) & 26.43 (29.55) \\ \hline
        5 & \textbf{11.34 (6.30)} & 25.22 (33.13) & 20.76 (29.36) \\ \hline
        7 & \textbf{6.03 (4.58)} & 25.27 (46.18) & 16.14 (35.25) \\ \hline
        Succ. & \textbf{39/50} & 12/50 & 14/50 \\ \hline
        \hline
        \multicolumn{4}{|c|}{\texttt{036\_drill\_scanned}} \\ \hline
        n & DDIM & SDF & SDF (FT) \\ \hline
        1 & \textbf{51.63 (26.70)} & 63.89 (23.12) & 59.70 (25.20) \\ \hline
        3 & \textbf{18.89 (21.29)} & 35.12 (43.13) & 27.07 (39.28) \\ \hline
        5 & \textbf{4.59 (6.09)} & 9.03 (48.61) & 10.97 (47.16) \\ \hline
        7 & \textbf{3.87 (4.19)} & 8.92 (54.90) & 13.37 (49.80) \\ \hline
        Succ. & \textbf{43/50} & 26/50 & 22/50 \\ \hline
    \end{tabular}
\end{table}

Results in Table~\ref{tab:rw_static} show consistent sim-to-real transfer. For geometrically simple objects (mustard bottle), all methods converge reliably with marginal differences. For objects with complex geometry (mug, power drill), local sampling baselines exhibit high variance and substantially lower success rates, indicating frequent failures to converge towards accurate estimates. High variance reflects insufficient coverage of valid hypotheses that can lead to convergence towards incorrect estimates. This degradation is particularly evident as the number of contacts increases, where baselines show persistent high ADD values despite additional observations. In contrast, the learned inverse model maintains consistent convergence behavior with lower variance and higher success rates. However, coverage of the hypotheses space is less uniform due to unmodeled sensor effects, which can lead to out-of-distribution (OOD) observations and mode collapse. An example run is shown in Figure~\ref{fig:mug_real}. Full runs and failure modes are provided in the supplementary video.

\subsection{Object Pose Tracking in Box Pushing}
\label{subsec:box_pushing}
This experiment evaluates robustness under contact dynamics not explicitly modeled during training, testing the framework under challenging conditions that combine unmodeled dynamics with partial observability from featureless geometry. The pushing scenario involves tracking the pose of a bulky box through sequences of pushing contacts with varying durations, including sticking, sliding, and contact breaks.

\subsubsection{Simulated Results}

The pushing scenario is implemented in MuJoCo using a heavy box pushed with constant end-effector velocity via teleoperation. We carry out 10 recorded pushing experiments with varying contact sequences, running estimation 5 times per recording to account for estimator variance. State transition updates in the particle filter are performed using multi-step dynamics rollouts on the existing particles.

\begin{table}[h]
    \centering
    \caption{Box pushing in simulation. Median and IQR of ADD ($\times 10^{-2}$) of the final belief, evaluated over 50 runs (5 reruns per 10 recordings). Values below 10 indicate successful convergence (ADD $< 0.1$).}
    \label{tab:sim_comparison}
    \begin{tabular}{|c|c|c|}
        \hline
        Method & ADD ($\times 10^{-2}$) & Success \\ \hline
        SDF & 24.35 (30.58) & 11/50 \\ \hline
        SDF (FT) & 18.12 (29.30) & 19/50 \\ \hline
        \textbf{DDIM} & \textbf{7.71 (5.66)} & \textbf{35/50} \\ \hline
    \end{tabular}
\end{table}

\begin{figure}[h]
  \centering
  \includegraphics[width=0.9\linewidth]{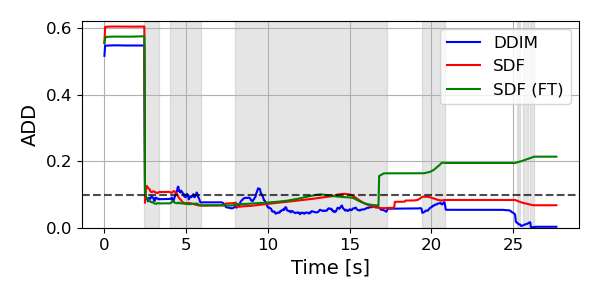}
  \caption{Example trajectory from simulated box pushing. ADD of the averaged belief over time comparing the learned inverse model against local sampling baselines. Shaded regions indicate contact periods and a dashed line marks the 0.1 accuracy threshold.}
  \label{fig:box_pushing_sim}
\end{figure}

Results in Table~\ref{tab:sim_comparison} show that for the featureless box geometry, local sampling baselines exhibit high variance and low success rates, reflecting inability to adequately cover the hypothesis space in a tracking scenario with limited geometric constraints. The learned inverse model maintains more consistent convergence behavior, with reduced variance and higher success rate. An example run is shown in Figure~\ref{fig:box_pushing_sim}. Computational performance averages $38.7 \pm 4.2$\,ms per iteration on a 2.80\,GHz Intel Core i7-1165G7.

\subsubsection{Real-World Validation}

Models trained purely on simulated data with domain randomization are further tested in a real-world box-pushing scenario. Data is collected via teleoperation with multiple pushing contacts involving sliding and contact breaks. Estimation is run 5 times for each of 8 recorded experiments to account for estimator variance.

\begin{table}[h]
    \centering
    \caption{Box pushing in real-world experiments. Median and IQR of ADD ($\times 10^{-2}$) of the final belief, evaluated over 40 runs (5 reruns per 8 recordings). Values below 10 indicate successful convergence (ADD $< 0.1$).}
    \label{tab:multi_seed_comparison}
    \begin{tabular}{|c|c|c|}
        \hline
        Method & ADD ($\times 10^{-2}$) & Success \\ \hline
        SDF & 14.39 (12.36) & 8/40 \\ \hline
        SDF (FT) & 16.05 (13.35) & 7/40 \\ \hline
        \textbf{DDIM} & \textbf{9.06 (5.27)} & \textbf{21/40} \\ \hline
    \end{tabular}
\end{table}

\begin{figure}[h]
  \centering
  \includegraphics[width=0.9\linewidth]{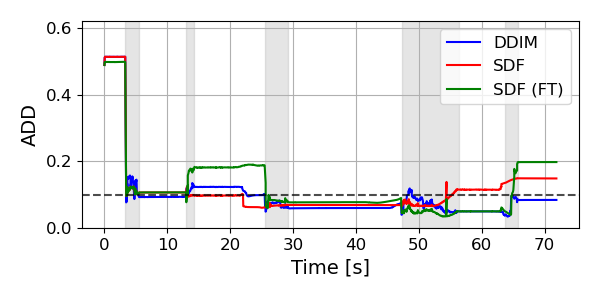}
  \caption{Example trajectory from real-world box pushing. ADD of the averaged belief over time comparing the learned inverse model against local sampling baselines. Shaded regions indicate contact periods, and the dashed line marks the 0.1 accuracy threshold.}
  \label{fig:rw_pushing}
\end{figure}

Real-world results in Table~\ref{tab:multi_seed_comparison} are consistent with the simulated experiments. Local sampling baselines fail to converge towards accurate estimates in the majority of runs, while the learned inverse model achieves a substantially higher accuracy and success rate due to tactile-conditioning. An example run is shown in Figure~\ref{fig:rw_pushing}, with qualitative results in Figure~\ref{fig:box_real_qualitative}. Full runs and failure modes are provided in the supplementary video.

\begin{figure}[h]
  \centering
  \includegraphics[width=0.9\linewidth]{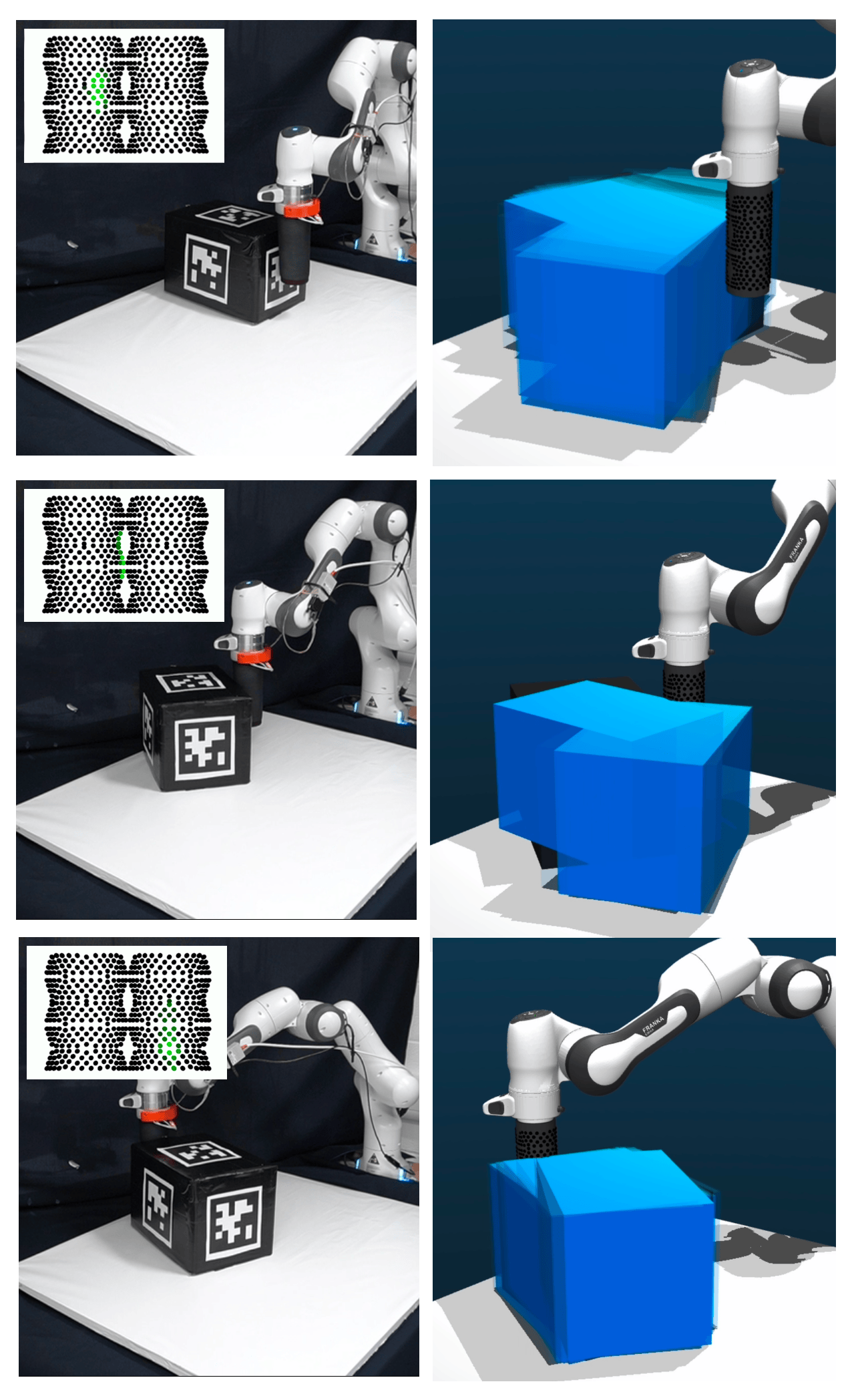}
  \caption{Real-world box pushing experiment. \textit{Left:} Sequential contacts with different box faces, with tactile activations shown in green. An AprilTag provides ground-truth pose for evaluation only. \textit{Right:} Particle filter belief visualization with opacity proportional to particle weight.}
  \label{fig:box_real_qualitative}
\end{figure}

\section{Discussion}
\label{sec:discussion}

\paragraph{Inverse Sensor Modeling.}

The learned inverse sensor model provides sample-efficient generation of object pose hypotheses from tactile observations, improving upon local sampling across all test objects (Figure~\ref{fig:radar_chart}). This translates to substantially lower hypothesis errors than baselines across sensor resolutions (Figure~\ref{fig:taxel_density}), with accuracy improving as observations become more discriminative. For distinctive contact geometries at higher resolutions, accurate hypotheses can be generated from a single contact. Additionally, as demonstrated in the box-pushing examples (Table~\ref{tab:multi_seed_comparison}), tactile-conditioned hypotheses can help recover from diverging beliefs and compensate for dynamics modeling errors. The simulated forward sensor model enables scalable dataset generation without real-world data collection, and the simple denoiser architecture allows fast training and online inference on consumer-grade CPU hardware (Table~\ref{tab:inference}).

\paragraph{Geometry-Aware Inference.}
The SDF-based projection step provides contact constraint satisfaction in generated pose hypotheses, ensuring physical validity and high-likelihood samples for downstream estimation. By incorporating geometric priors, it offloads constraint enforcement from the learning process, reducing training requirements and allowing for the use of smaller denoiser architectures. Ablation experiments confirm that removing constraint enforcement decreases performance below local sampling baselines (Figure~\ref{fig:radar_chart}), although accurate pose hypotheses are still generated due to tactile conditioning (Figure~\ref{fig:taxel_density}). Since the diffusion model learns to generate samples near the contact manifold, projection requires only small corrections and adds negligible computational overhead when combined with SDF precomputation (Table~\ref{tab:inference}).

\paragraph{Belief-Informed Particle Injection.}
Combining the learned inverse sensor model with a particle filter further offloads the learning process by utilizing the inverse model for particle proposal, and a standard Bayesian estimation framework for belief tracking over sequential contacts. Additionally, it provides an interpretable framework that allows for integration of model-based or learned dynamics and observation models. The proposed particle injection method enables pose estimation from sequential contacts with faster convergence and higher accuracy than local sampling baselines (Figure~\ref{fig:planar_estim}), effectively addressing particle depletion under partial observability. As the generated hypotheses cover multimodal distributions on the contact manifold, the framework does not depend on tight initial pose priors. It is capable of maintaining multimodal estimates, while integrating prior beliefs and preserving plausible modes (Figure~\ref{fig:intro}). Notably, simulated force-torque observations sometimes outperform full tactile observations when using local sampling, demonstrating that generating high-likelihood hypotheses is critical for leveraging the discriminative observations provided by the tactile skin. These results translate to real-world static pose estimation, where the proposed method achieves higher accuracy and success rates than baselines (Table~\ref{tab:rw_static}), although higher variance and mode collapse are observed due to imperfect contact and unmodeled dynamics (Figure~\ref{fig:mug_real}). Tracking experiments further demonstrate robustness to pushing dynamics not modeled during training (Figure~\ref{fig:box_real_qualitative}), with efficient hypothesis generation enabling online filtering at over 20\,Hz on CPU.

\subsection{Limitations and Future Work}

The planar setting constrains estimation to 3-DoF, and extension to 6-DoF would increase both data requirements and computational cost, likely necessitating GPU acceleration. Furthermore, our current implementation assumes a cylindrical end-effector, which simplifies contact generation. Complex contact geometries would require more sophisticated projection methods with added computational overhead.
Additionally, the geometric sensor model used for training does not consider dynamic effects during sliding or impact. Capturing these effects would require more extensive domain priors through physics-based modeling or real-world training data. Training is also object-specific and assumes known geometries, which could be further addressed through category-level learning or by learning to jointly infer geometry and pose. Furthermore, generative models may inherit training distribution biases, leading to mode collapse or particle depletion in downstream estimators. This makes training data diversity critical for generalization. The utilized SDF-based projection method may introduce bias in contact manifold coverage, which we address partially through subsampling and binning, though exploration-based methods might offer a promising alternative for guaranteeing uniform coverage. In some cases, premature mode collapse can also be caused by the belief-informed injection approach, which is sensitive to tuning and could benefit from explicitly tracking sample diversity.
In addition to addressing these limitations, future directions include active exploration strategies for selecting contacts that maximize information gain, as well as extending the framework to handle multiple simultaneous contacts for whole-body and cluttered scenarios.

\section{Conclusion}
\label{sec:conclusion}
We introduced a diffusion-based inverse sensor model integrated with particle filtering for vision-free object pose estimation using a distributed tactile sensor. The approach combines learned generative modeling with geometric priors through SDF-based contact simulation and constraint enforcement, enabling data-efficient learning across sensor resolutions and object geometries. Experiments in simulated and real-world planar scenarios demonstrate improved sample efficiency, faster convergence, and higher accuracy compared to local sampling baselines, while maintaining multimodal beliefs and exhibiting robustness to unmodeled dynamics through training-time randomization. These results suggest that integrating learned generative models with explicit geometric reasoning and particle filtering provides a principled pathway toward robust tactile state estimation.
Current limitations include object-specific training, restriction to planar scenarios, and reliance on known geometry. Future work will consider extensions to 6-DoF settings, multi-object scenarios with simultaneous contacts, and joint learning of object geometry.

\begin{dci}
The author(s) declared no potential conflicts of interest with respect to the research, authorship, and/or publication of this article.  
\end{dci}

\begin{funding}
This work was supported by the State Secretariat for Education, Research and Innovation in Switzerland for participation in the European Commission's Horizon Europe Program through the SESTOSENSO project (\url{https://sestosenso.eu/}) and the INTELLIMAN project (\url{https://intelliman-project.eu/}).
\end{funding}

\bibliographystyle{SageH}
\bibliography{main.bib}

\vfill\eject
\setcounter{secnumdepth}{2}
\appendix
\renewcommand{\thesection}{\Alph{section}} 

\section{Implementation Details and Parameters}
\label{app:implementation}

\subsection{Particle Filter Parameters}
\label{app:pf_resampling}
\label{app:lvr}

\begin{table}[h]
    \centering
    \footnotesize
    \caption{Particle filter and workspace parameters.}
    \label{tab:pf_params}
    \begin{tabular}{ccc}
    \toprule
        Parameter & Symbol & Value \\
    \midrule
        Number of particles & $N$ & $100-300$ \\
        Injected particles per step & $N_p$ & $100-300$ \\
        ESS resampling threshold & $ESS_{\max}$ & $0.6N$ \\
        Kernel bandwidth bounds & $[h_{\min}, h_{\max}]$ & $[0.02, 0.1]$ \\
        $k$-NN for consistency & $k$ & $5$ \\
        Observation noise range & $[\tilde{\sigma}_{\min}, \tilde{\sigma}_{\max}]$ & $[0.4, 1.2]$ \\
        Likelihood transition sharpness & $\kappa$ & 1000 \\
        Likelihood transition offset & $d_0$ & 0.01\,m \\    \midrule
        \multicolumn{3}{c}{\textit{Workspace bounds}} \\
    \midrule
        $x$ position & - & $[0.2, 0.6]$ m \\
        $y$ position & - & $[-0.3, 0.3]$ m \\
        $\theta$ orientation & - & $[0, 2\pi]$ rad \\
        $\theta$ (symmetric objects) & - & $[0, \pi]$ rad \\
    \bottomrule
    \end{tabular}
\end{table}

\FloatBarrier
\subsection{Baseline Implementations}
\label{app:baseline}

\begin{algorithm}[h]
\caption{Local sampling baseline}
\footnotesize
\label{alg:sdfproposer_short}
\KwIn{Predicted belief $\bar{\mathcal{X}}_{t}$, observation $\mathbf{z}_t$, sensor pose $\mathbf{u}_t$}
\KwOut{Updated belief $\mathcal{X}_t$}

\ForEach{particle $\langle \bar{\mathbf{x}}^o, \bar{w}_{t} \rangle \in \bar{\mathcal{X}}_t$}{
$\bar{w}_t \leftarrow \bar{w}_{t}\,
\mathcal{L}_{\text{obs}}(\mathbf{z}_t \mid {}^s\bar{\mathbf{x}}^{o}_t)
$\;
}

$\hat{\mathcal{X}}_t \leftarrow \textsc{LVR}(\bar{\mathcal{X}}_{t})$\;
\ForEach{particle $\langle \hat{\mathbf{x}}^o, \hat{w}_{t} \rangle \in \hat{\mathcal{X}}_t$}{
$\hat{\mathbf{x}}^{o,[i]}_t \sim \hat{\mathbf{x}}^{o,[i]}_t + \boldsymbol{\epsilon},
\ \boldsymbol{\epsilon} \sim \mathcal{U}(\boldsymbol{\sigma}_{\min}, \boldsymbol{\sigma}_{\max})$\;
$\hat{\mathbf{x}}^{o,[i]}_t \leftarrow
\text{project}_{\phi,\nabla\phi}(\hat{\mathbf{x}}^{o,[i]}_t;\, \mathbf{u}_t)$\;
$\ell_t \leftarrow \log\mathcal{L}_{\mathrm{obs}}(\mathbf{z}_t \mid {}^s\hat{\mathbf{x}}^o)
+ \tilde{\ell}_t(\hat{\mathbf{x}}^o)$\;
$\hat{w}_t \leftarrow \exp(\ell_t)$\;
}

$\mathcal{Y}_t \leftarrow \bar{\mathcal{X}}_t \cup \hat{\mathcal{X}}_t$\;
$\mathcal{X}_t \leftarrow \textsc{LVR}(\mathcal{Y}_t)$\;
\Return{$\mathcal{X}_t$}
\end{algorithm}

\begin{table}[h]
    \centering
    \footnotesize
    \caption{Local sampling baseline parameters.}
    \label{tab:baseline_params}
    \begin{tabular}{ccc}
    \toprule
        Parameter & Symbol & Value \\
    \midrule
         Radius range & $[\sigma_{r_{\min}}, \sigma_{r_{\max}}]$ & $[0, 0.03]$ \\
         Angle range & $[\sigma_{\alpha_{\min}}, \sigma_{\alpha_{\max}}]$ & $[-\pi, \pi]$ \\
         Orientation range & $[\sigma_{\theta_{\min}}, \sigma_{\theta_{\max}}]$ & $[-\pi, \pi]$ \\
         Orientation scale & $k_\theta$ & $0.6$ \\
         SDF projection tol. & $\delta$ & $0.003$ \\
    \bottomrule
    \end{tabular}
\end{table}

The local sampling baseline is detailed in Algorithm~\ref{alg:sdfproposer_short}. Perturbations are sampled using cylindrical coordinates, with bounds $\boldsymbol{\sigma}_n = [\sigma_{r}, \sigma_{\alpha}, k_\theta^{n-1}\sigma_{\theta}]$ progressively tightened with each contact $n$. They are additionally clipped to $0.1$ to retain stochastic sampling at higher $n$ values. SDF projection ensures all hypotheses satisfy contact constraints (Section~\ref{sec:contact_generation}). Parameters are provided in Table~\ref{tab:baseline_params}.

\paragraph{Force-torque baseline.}
The simulated force-torque baseline condenses the distributed tactile response into a single virtual contact point. Taxels with activations above a noise threshold $\zeta$ are averaged to obtain a mean activation $\bar{z}$ and centroid position $\bar{\mathbf{q}}$, which is projected onto the sensor surface. Particle likelihoods are evaluated as $\mathcal{L}^{[n]} = \tilde{p}(\bar{z} \mid {}^o\bar{\mathbf{q}})$, reducing the observation to a scalar at a single contact point.

\subsection{Diffusion Model Training and Inference}
\label{app:diffusion}

\paragraph{Architecture.} The denoiser $\boldsymbol{\epsilon}_\theta$ is a multi-layer perceptron (MLP) with three fully connected layers of 128 units each and ReLU activations. It takes as input a noisy pose ${}^s\mathbf{x}^o_t \in \mathbb{R}^3$, timestep $t \in \{1, \dots, T\}$, and tactile observations $\mathbf{z} \in [0,1]^{N_\text{tax}}$, and outputs predicted noise $\hat{\boldsymbol{\epsilon}} \in \mathbb{R}^3$.

\paragraph{DDPM forward diffusion step}
As in~\cite{ho_denoising_2020}, the forward diffusion steps add Gaussian noise to ground truth poses over $T$ timesteps. Noisy poses at step $t$ are obtained as
\begin{align}
{}^s\mathbf{x}^{o}_{t}
=
\sqrt{\bar{\alpha}_t}\,{}^s\mathbf{x}^{o}
+
\sqrt{1-\bar{\alpha}_t}\,\boldsymbol{\epsilon}_{t},
\end{align}
with sampled noise $\boldsymbol{\epsilon}_{t}\sim\mathcal{N}(0,\mathbf{I})$ and cumulative noise schedule $\bar{\alpha}_t$.

\paragraph{DDIM inference step}
Given a tactile observation $\mathbf{z}$, latent pose
${}^s\mathbf{x}^o_{\tau_S} \sim \mathcal{N}(0,\mathbf{I})$
is iteratively denoised over $S$ timesteps
$\tau=\{\tau_1, \dots, \tau_S\}$ with $\tau_S=T$ and $\tau_1=1$.
At each step, the denoiser predicts the added noise conditioned on the tactile observation
$\hat{\boldsymbol{\epsilon}} = \boldsymbol{\epsilon}_\theta({}^s\mathbf{x}^o_{\tau_i},\,\mathbf{z},\,\tau_i)$,
and updates the pose hypotheses using the DDIM update~\citep{song_denoising_2022}
\begin{align}
{}^s\mathbf{x}^o_{\tau_{i-1}}
&=
\sqrt{\bar{\alpha}_{\tau_{i-1}}}\,\hat{\mathbf{x}}_0
+
\sqrt{1-\bar{\alpha}_{\tau_{i-1}} - \sigma_{\tau_i}^2}\,\hat{\boldsymbol{\epsilon}}
+
\sigma_{\tau_i}\,\boldsymbol{\omega},
\end{align}
where
$\hat{\mathbf{x}}_0 = \bigl({}^s\mathbf{x}^o_{\tau_i} - \sqrt{1-\bar{\alpha}_{\tau_i}}\,\hat{\boldsymbol{\epsilon}}\bigr) / \sqrt{\bar{\alpha}_{\tau_i}}$
is the predicted denoised sample, $\boldsymbol{\omega}\!\sim\!\mathcal{N}(0,\mathbf{I})$, and $\eta\!\in\![0,1]$ controls the stochasticity through
\begin{align}
\sigma_{\tau_i}
=
\eta
\sqrt{
\frac{1-\bar{\alpha}_{\tau_{i-1}}}{1-\bar{\alpha}_{\tau_i}}
\left(
1 - \frac{\bar{\alpha}_{\tau_i}}{\bar{\alpha}_{\tau_{i-1}}}
\right)
}.
\end{align}
Parameters are provided in Table~\ref{tab:diffusion_params}.
\begin{table}[h]
    \centering
    \footnotesize
    \caption{Diffusion model parameters.}
    \label{tab:diffusion_params}
    \begin{tabular}{ccc}
    \toprule
        Parameter & Symbol & Value \\
    \midrule
        \multicolumn{3}{c}{\textit{Training}} \\
    \midrule
        Dataset size & $N_d$ & $10^5$ \\
        Epochs & - & $3000$ \\
        Batch size & - & $64$ \\
        Optimizer & - & Adam \\
        Learning rate & - & $10^{-3}$ \\
        Step size & - & $100$ \\
        Gamma & - & $0.95$ \\
        Early stopping & - & $200$ epochs \\
        Loss scaling & $\text{diag}(\boldsymbol{\lambda})$ & $[1, 1, 0.1]$ \\
    \midrule
        \multicolumn{3}{c}{\textit{Diffusion process}} \\
    \midrule
        Number of steps & $T$ & $100$ \\
        Noise schedule & - & Linear \\
    \midrule
        \multicolumn{3}{c}{\textit{DDIM inference}} \\
    \midrule
        Sampling steps & $S$ & $80$ \\
        Stochasticity & $\eta$ & $0.2$ \\
        SDF projection tol. & $\delta$ & $0.003$ \\
    \bottomrule
    \end{tabular}
\end{table}

\paragraph{Dataset generation.} Training data is generated by uniform sampling from $\mathcal{X}^o$ (Table \ref{tab:pf_params}) followed by SDF projection. To compensate for non-uniform density caused by projection to convex surfaces, histogram-based binning is applied. Poses are binned jointly over contact angle ($n_{\text{pos}}=50$ bins) and object orientation ($n_{\text{ori}}=100$ bins), with $n_{\text{bin}}=10$ samples retained per bin.

\section{Hardware and Sensor Details}
\label{app:hardware}
The sensing end-effector uses CySkin capacitive tactile technology~\citep{maggiali2008embedded}, consisting of interconnected triangular flexible PCB modules, each containing 10 taxels, arranged in a mesh structure that conforms to curved surfaces. A conductive Lycra layer serves as the deformable electrode, wrapped over the end-effector halves. A 3~mm silicone elastomer layer enables conforming to local object geometry and acts as a spatial pressure filter. A fabric sleeve forms the protective outer layer. Table~\ref{tab:sensor_specs} summarizes key sensor specifications, and parameters of the geometric sensor model used in simulation are listed in Table~\ref{tab:sensor_model_params}.
\begin{table}[h]
    \centering
    \footnotesize
    \caption{CySkin sensor and communication specifications.}
    \label{tab:sensor_specs}
    \begin{tabular}{cc}
    \toprule
        Parameter & Value \\
    \midrule
        Taxel diameter / pitch & $3.5$ mm / $8$ mm \\
        Number of taxels & $514$ (non-homogeneous) \\
        End-effector dimensions & $r = 0.035$ m, $h = 0.2$ m \\
        Sensing surface height & $0.15$ m \\
        Sampling rate & $20$ Hz \\
        Communication & SPI $\to$ IHB $\to$ CAN bus \\
        Normalization constant & $z_{\max} = 3000$ \\
        Noise threshold & $\zeta = 0.2$ \\
    \bottomrule
    \end{tabular}
\end{table}

\paragraph{Signal Preprocessing.}
Quantized capacitance readings are processed by first subtracting a baseline offset recorded at initialization (no contact), then thresholding negative activations to zero, and finally normalizing to $[0,1]$ using calibration values. Specifically, each taxel reading is computed as $z^i = z^i_{\text{raw}} - z^i_{\text{baseline}}$ and clipped to zero if negative, then normalized as $\tilde{z}^i = z^i / z_{\max}$. Finally, noise is filtered by clipping $z < \zeta$ to $0$. Taxel positions $\{{}^s\mathbf{q}^i\}_{i=1}^{N_\text{tax}}$ are obtained through spatial calibration~\citep{albini2017towards}.

\begin{table}[h]
    \centering
    \footnotesize
    \caption{Tactile observation model parameters.}
    \label{tab:sensor_model_params}
    \begin{tabular}{ccc}
    \toprule
        Parameter & Symbol & Value \\
    \midrule
        Observation noise std. dev. & $\sigma_{tax}$ & $0.02$ \\
        Max activation distance & $d_{max}$ & $0.003$ m \\
        Penetration threshold & $\delta_{\text{pen}}$ & $-0.003$ m \\
        Max distance threshold & $\delta_{\text{max}}$ & $0.0$ m \\
        Inactive patch prob. & $p_{fail}$ & $0$ (sim) / $0.7$ (real) \\
    \bottomrule
    \end{tabular}
\end{table}

\section{Object Models and Geometry Processing}
\label{app:objects}

\begin{table}[h]
    \centering
    \footnotesize
    \caption{List of utilized test objects and parameters. The ADD-S metric is used for symmetric objects, and ADD for others. End-effector height $z_{ee}$ adjusted per object.}
    \label{tab:object_list}
    \footnotesize
    \begin{tabular}{ccccc}
    \toprule
        Object ID & Name & Symmetry & $z_{ee}$ (m) & $d_{obj}$ (m)\\
    \midrule
        001 & Bulky box & Discrete & $0.30$ & $0.4573$\\
        002 & Master chef can & Continuous & $0.20$ & $0.1720$\\
        003 & Cracker box & Discrete & $0.20$ & $0.2695$\\
        006 & Mustard bottle & Discrete & $0.20$ & $0.1965$\\
        019 & Pitcher base & None & $0.30$ & $0.2595$\\
        024 & Bowl & Continuous & $0.20$ & $0.1620$\\
        025 & Mug & None & $0.20$ & $0.1250$\\
        035 & Power drill & None & $0.18$ & $0.2263$\\
        036 & Scanned drill & None & $0.18$ & $0.2388$\\
        061 & Foam brick & Discrete & $0.20$ & $0.1030$\\
        077 & Rubik's cube & Discrete & $0.20$ & $0.0956$\\
    \bottomrule
    \end{tabular}
\end{table}
Object meshes are sourced from the YCB dataset~\citep{YCBds}, aside from \texttt{001\_bulky\_box} which uses a custom mesh. Additionally, real-world experiments utilized \texttt{036\_power\_drill}, which is reconstructed from scanned point cloud data of a lab unit that differs from the YCB model and can be seen in the supplementary video. For symmetry handling, ADD-S is used for both discrete and continuous symmetries.

\paragraph{SDF Computation.}
SDFs are precomputed per object by constructing a bounding box of size $0.4~m\times0.4~m\times0.3~m$, discretizing it into a $128^3$ voxel grid, computing ray-cast signed distances, and storing the resulting field and gradients. Runtime distance queries use trilinear interpolation. Gradients are computed numerically using central differences.

\section{Additional Quantitative Results}
\label{app:results}

\subsection{Sample Efficiency Analysis}
\label{app:sampling_efficiency}
\label{app:sampling_diagnostics}
Tables~\ref{tab:taxels-a-003}-\ref{tab:taxels-a-035} show the median and IQR of displacement errors (ADD$\times 10^{-2}$) of MAP estimates across sensor resolutions (fixed $N=100$). Tables~\ref{tab:taxels-b-003}-\ref{tab:taxels-b-035} show results across sample sizes (fixed resolution 1.56), corresponding to the analysis in Section~\ref{subsec:sample_efficiency}.

\begin{table}[h]
    \centering
    \footnotesize
    \caption{Taxel density ablation: \texttt{003\_cracker\_box}, $N=100$ samples.}
    \label{tab:taxels-a-003}
    \begin{tabular}{|c|c|c|c|}
        \hline
        \multicolumn{4}{|c|}{\texttt{003\_cracker\_box}} \\ \hline
        Res & DDIM & DDIM (no SDF) & SDF \\ \hline
        0.29 & \textbf{1.64 (1.23)} & 2.55 (1.10) & 3.75 (1.63) \\ \hline
        0.79 & \textbf{1.11 (0.63)} & 2.18 (0.95) & 3.81 (1.62) \\ \hline
        1.56 & \textbf{0.89 (0.54)} & 1.79 (0.73) & 3.81 (1.66) \\ \hline
        2.3 & \textbf{0.81 (0.55)} & 1.89 (1.01) & 3.76 (1.65) \\ \hline
    \end{tabular}
\end{table}

\begin{table}[h]
    \centering
    \footnotesize
    \caption{\texttt{006\_mustard\_bottle}, $N=100$ samples.}
    \label{tab:taxels-a-006}
    \begin{tabular}{|c|c|c|c|}
        \hline
        \multicolumn{4}{|c|}{\texttt{006\_mustard\_bottle}} \\ \hline
        Res & DDIM & DDIM (no SDF) & SDF \\ \hline
        0.29 & \textbf{1.32 (0.79)} & 2.33 (1.07) & 3.06 (1.40) \\ \hline
        0.79 & \textbf{0.89 (0.53)} & 1.90 (0.95) & 3.02 (1.37) \\ \hline
        1.56 & \textbf{0.70 (0.39)} & 1.84 (0.80) & 3.04 (1.28) \\ \hline
        2.3 & \textbf{0.89 (0.45)} & 1.84 (0.76) & 3.04 (1.31) \\ \hline
    \end{tabular}
\end{table}

\begin{table}[h]
    \centering
    \footnotesize
    \caption{\texttt{025\_mug}, $N=100$ samples.}
    \label{tab:taxels-a-025}
    \begin{tabular}{|c|c|c|c|}
        \hline
        \multicolumn{4}{|c|}{\texttt{025\_mug}} \\ \hline
        Res & DDIM & DDIM (no SDF) & SDF \\ \hline
        0.29 & \textbf{6.76 (8.96)} & 11.19 (11.25) & 13.30 (9.55) \\ \hline
        0.79 & \textbf{3.74 (2.58)} & 7.50 (4.17) & 13.32 (9.25) \\ \hline
        1.56 & \textbf{3.31 (2.97)} & 6.85 (5.10) & 12.86 (9.35) \\ \hline
        2.3 & \textbf{3.07 (2.42)} & 7.89 (5.22) & 13.07 (9.68) \\ \hline
    \end{tabular}
\end{table}

\begin{table}[h]
    \centering
    \footnotesize
    \caption{\texttt{035\_power\_drill}, $N=100$ samples.}
    \label{tab:taxels-a-035}
    \begin{tabular}{|c|c|c|c|}
        \hline
        \multicolumn{4}{|c|}{\texttt{035\_power\_drill}} \\ \hline
        Res & DDIM & DDIM (no SDF) & SDF \\ \hline
        0.29 & \textbf{2.62 (2.10)} & 4.37 (2.68) & 15.23 (10.88) \\ \hline
        0.79 & \textbf{2.57 (1.85)} & 3.69 (2.13) & 15.43 (11.38) \\ \hline
        1.56 & \textbf{1.94 (1.78)} & 3.74 (2.57) & 15.22 (10.86) \\ \hline
        2.3 & \textbf{1.80 (1.24)} & 3.46 (1.92) & 15.27 (10.74) \\ \hline
    \end{tabular}
\end{table}

\begin{table}[h]
    \centering
    \footnotesize
    \caption{\texttt{003\_cracker\_box}, resolution 1.56.}
    \label{tab:taxels-b-003}
    \begin{tabular}{|c|c|c|c|}
        \hline
        \multicolumn{4}{|c|}{\texttt{003\_cracker\_box}} \\ \hline
        Samples & DDIM & DDIM (no SDF) & SDF \\ \hline
        1 & \textbf{5.27 (4.50)} & 7.73 (4.35) & 37.51 (29.24) \\ \hline
        10 & \textbf{1.97 (1.13)} & 3.25 (1.59) & 8.76 (4.43) \\ \hline
        100 & \textbf{0.89 (0.54)} & 1.79 (0.73) & 3.81 (1.66) \\ \hline
        1000 & \textbf{0.47 (0.35)} & 0.96 (0.49) & 2.05 (1.10) \\ \hline
    \end{tabular}
\end{table}

\begin{table}[h]
    \centering
    \footnotesize
    \caption{\texttt{006\_mustard\_bottle}, resolution 1.56.}
    \label{tab:taxels-b-006}
    \begin{tabular}{|c|c|c|c|}
        \hline
        \multicolumn{4}{|c|}{\texttt{006\_mustard\_bottle}} \\ \hline
        Samples & DDIM & DDIM (no SDF) & SDF \\ \hline
        1 & \textbf{6.24 (7.20)} & 10.00 (6.72) & 41.68 (31.32) \\ \hline
        10 & \textbf{1.61 (1.24)} & 3.85 (2.28) & 6.27 (4.50) \\ \hline
        100 & \textbf{0.70 (0.39)} & 1.84 (0.80) & 3.04 (1.28) \\ \hline
        1000 & \textbf{0.47 (0.34)} & 0.92 (0.41) & 1.57 (0.65) \\ \hline
    \end{tabular}
\end{table}

\begin{table}[h]
    \centering
    \footnotesize
    \caption{\texttt{025\_mug}, resolution 1.56.}
    \label{tab:taxels-b-025}
    \begin{tabular}{|c|c|c|c|}
        \hline
        \multicolumn{4}{|c|}{\texttt{025\_mug}} \\ \hline
        Samples & DDIM & DDIM (no SDF) & SDF \\ \hline
        1 & \textbf{39.35 (34.40)} & 41.45 (27.10) & 106.02 (55.02) \\ \hline
        10 & \textbf{9.25 (8.36)} & 17.58 (11.52) & 38.01 (26.72) \\ \hline
        100 & \textbf{3.31 (2.97)} & 6.85 (5.10) & 12.86 (9.35) \\ \hline
        1000 & \textbf{1.69 (0.99)} & 3.29 (2.56) & 4.75 (2.95) \\ \hline
    \end{tabular}
\end{table}

\begin{table}[h]
    \centering
    \footnotesize
    \caption{\texttt{035\_power\_drill}, resolution 1.56.}
    \label{tab:taxels-b-035}
    \begin{tabular}{|c|c|c|c|}
        \hline
        \multicolumn{4}{|c|}{\texttt{035\_power\_drill}} \\ \hline
        Samples & DDIM & DDIM (no SDF) & SDF \\ \hline
        1 & 22.94 (16.82) & \textbf{19.78 (13.10)} & 94.45 (45.54) \\ \hline
        10 & \textbf{7.92 (4.18)} & 8.61 (4.78) & 44.71 (19.91) \\ \hline
        100 & \textbf{1.94 (1.78)} & 3.74 (2.57) & 15.22 (10.86) \\ \hline
        1000 & \textbf{1.12 (0.84)} & 1.79 (1.10) & 4.62 (4.63) \\ \hline
    \end{tabular}
\end{table}

\subsection{Multi-Contact Pose Estimation}
\label{app:pose_estimation}
\label{app:full_results}

Tables~\ref{tab:001_bulky_box}--\ref{tab:077_rubiks_cube} show the median and IQR of displacement errors (ADD$\times 10^{-2}$) of averaged beliefs vs.\ contact count $n$, and the success rate (ADD $< 0.1$) for all test objects.

\begin{table}[h]
    \centering
    \footnotesize
    \caption{\texttt{001\_bulky\_box}}
    \label{tab:001_bulky_box}
    \begin{tabular}{|c|c|c|c|}
    \hline
    \multicolumn{4}{|c|}{\texttt{001\_bulky\_box}} \\ \hline
    n & DDIM & SDF & SDF (FT) \\ \hline
    1 & \textbf{8.41 (3.73)} & 10.80 (4.54) & 10.10 (4.45) \\ \hline
    3 & \textbf{4.17 (3.59)} & 6.78 (4.81) & 6.34 (4.61) \\ \hline
    5 & \textbf{2.32 (2.07)} & 4.50 (5.81) & 3.94 (5.53) \\ \hline
    6 & \textbf{2.21 (1.76)} & 3.10 (5.72) & 2.51 (5.84) \\ \hline
    Succ. & \textbf{98/100} & 93/100 & 96/100 \\ \hline
    \end{tabular}
\end{table}

\begin{table}[h]
    \centering
    \footnotesize
    \caption{\texttt{002\_master\_chef\_can}}
    \label{tab:002_master_chef_can}
    \begin{tabular}{|c|c|c|c|}
    \hline
    \multicolumn{4}{|c|}{\texttt{002\_master\_chef\_can}} \\ \hline
    n & DDIM & SDF & SDF (FT) \\ \hline
    1 & \textbf{3.15 (1.40)} & 3.60 (2.20) & 3.94 (2.75) \\ \hline
    3 & \textbf{2.46 (0.40)} & 2.56 (0.41) & 2.60 (0.62) \\ \hline
    5 & \textbf{2.38 (0.74)} & 2.46 (0.63) & 2.47 (0.47) \\ \hline
    6 & \textbf{2.32 (0.67)} & 2.47 (0.42) & 2.47 (0.37) \\ \hline
    Succ. & \textbf{100/100} & 99/100 & 98/100 \\ \hline
    \end{tabular}
\end{table}

\begin{table}[h]
    \centering
    \footnotesize
    \caption{\texttt{003\_cracker\_box}}
    \label{tab:003_cracker_box}
    \begin{tabular}{|c|c|c|c|}
    \hline
    \multicolumn{4}{|c|}{\texttt{003\_cracker\_box}} \\ \hline
    n & DDIM & SDF & SDF (FT) \\ \hline
    1 & \textbf{5.33 (10.04)} & 11.35 (6.65) & 12.00 (5.19) \\ \hline
    3 & \textbf{2.38 (7.36)} & 3.80 (6.61) & 3.76 (6.44) \\ \hline
    5 & \textbf{2.05 (4.57)} & 2.54 (5.43) & 2.57 (4.05) \\ \hline
    6 & \textbf{1.83 (4.43)} & 2.41 (2.96) & 2.40 (2.62) \\ \hline
    Succ. & \textbf{88/100} & 84/100 & 87/100 \\ \hline
    \end{tabular}
\end{table}

\begin{table}[h]
    \centering
    \footnotesize
    \caption{\texttt{006\_mustard\_bottle}}
    \label{tab:006_mustard_bottle}
    \begin{tabular}{|c|c|c|c|}
    \hline
    \multicolumn{4}{|c|}{\texttt{006\_mustard\_bottle}} \\ \hline
    n & DDIM & SDF & SDF (FT) \\ \hline
    1 & \textbf{2.87 (3.47)} & 5.85 (2.99) & 6.07 (3.15) \\ \hline
    3 & \textbf{1.96 (1.13)} & 2.69 (1.20) & 2.71 (1.30) \\ \hline
    5 & \textbf{1.85 (0.85)} & 1.91 (0.84) & 1.97 (0.69) \\ \hline
    6 & 1.80 (0.92) & \textbf{1.72 (0.86)} & 1.87 (0.85) \\ \hline
    Succ. & \textbf{100/100} & \textbf{100/100} & \textbf{100/100} \\ \hline
    \end{tabular}
\end{table}

\begin{table}[h]
    \centering
    \footnotesize
    \caption{\texttt{019\_pitcher\_base}}
    \label{tab:019_pitcher_base}
    \begin{tabular}{|c|c|c|c|}
    \hline
    \multicolumn{4}{|c|}{\texttt{019\_pitcher\_base}} \\ \hline
    n & DDIM & SDF & SDF (FT) \\ \hline
    1 & \textbf{29.94 (21.63)} & 39.27 (39.57) & 38.40 (41.91) \\ \hline
    3 & \textbf{12.19 (18.32)} & 33.86 (27.03) & 29.48 (24.25) \\ \hline
    5 & \textbf{7.11 (13.06)} & 34.45 (37.03) & 20.99 (35.07) \\ \hline
    6 & \textbf{4.97 (10.13)} & 30.28 (39.34) & 19.62 (39.14) \\ \hline
    Succ. & \textbf{68/100} & 33/100 & 38/100 \\ \hline
    \end{tabular}
\end{table}

\begin{table}[h]
    \centering
    \footnotesize
    \caption{\texttt{024\_bowl}}
    \label{tab:024_bowl}
    \begin{tabular}{|c|c|c|c|}
    \hline
    \multicolumn{4}{|c|}{\texttt{024\_bowl}} \\ \hline
    n & DDIM & SDF & SDF (FT) \\ \hline
    1 & \textbf{9.65 (18.14)} & 45.20 (40.78) & 48.64 (37.72) \\ \hline
    3 & \textbf{3.08 (1.67)} & 8.79 (23.74) & 10.31 (26.80) \\ \hline
    5 & \textbf{2.64 (0.70)} & 3.30 (2.25) & 3.76 (3.10) \\ \hline
    6 & \textbf{2.53 (0.56)} & 2.91 (1.91) & 2.90 (1.72) \\ \hline
    Succ. & \textbf{99/100} & 92/100 & 90/100 \\ \hline
    \end{tabular}
\end{table}

\begin{table}[h]
    \centering
    \footnotesize
    \caption{\texttt{025\_mug}}
    \label{tab:025_mug}
    \begin{tabular}{|c|c|c|c|}
    \hline
    \multicolumn{4}{|c|}{\texttt{025\_mug}} \\ \hline
    n & DDIM & SDF & SDF (FT) \\ \hline
    1 & \textbf{50.28 (50.03)} & 74.61 (94.83) & 78.41 (100.77) \\ \hline
    3 & \textbf{19.35 (26.57)} & 45.55 (35.06) & 43.60 (34.50) \\ \hline
    5 & \textbf{7.64 (15.34)} & 26.37 (46.96) & 29.16 (46.57) \\ \hline
    6 & \textbf{5.47 (8.86)} & 22.21 (46.60) & 24.21 (42.76) \\ \hline
    Succ. & \textbf{72/100} & 45/100 & 37/100 \\ \hline
    \end{tabular}
\end{table}

\begin{table}[h]
    \centering
    \footnotesize
    \caption{\texttt{035\_power\_drill}}
    \label{tab:035_power_drill}
    \begin{tabular}{|c|c|c|c|}
    \hline
    \multicolumn{4}{|c|}{\texttt{035\_power\_drill}} \\ \hline
    n & DDIM & SDF & SDF (FT) \\ \hline
    1 & \textbf{17.33 (24.41)} & 45.82 (21.03) & 47.18 (24.58) \\ \hline
    3 & \textbf{4.39 (4.06)} & 37.73 (53.59) & 26.78 (44.27) \\ \hline
    5 & \textbf{2.94 (2.27)} & 28.97 (59.81) & 7.90 (46.51) \\ \hline
    6 & \textbf{2.42 (1.86)} & 13.61 (55.35) & 4.96 (48.06) \\ \hline
    Succ. & \textbf{96/100} & 50/100 & 57/100 \\ \hline
    \end{tabular}
\end{table}

\begin{table}[h]
    \centering
    \footnotesize
    \caption{\texttt{061\_foam\_brick}}
    \label{tab:061_foam_brick}
    \begin{tabular}{|c|c|c|c|}
    \hline
    \multicolumn{4}{|c|}{\texttt{061\_foam\_brick}} \\ \hline
    n & DDIM & SDF & SDF (FT) \\ \hline
    1 & \textbf{11.00 (8.37)} & 13.36 (23.51) & 12.35 (15.84) \\ \hline
    3 & \textbf{4.87 (3.90)} & 6.48 (4.38) & 6.70 (3.22) \\ \hline
    5 & \textbf{3.53 (2.40)} & 4.04 (3.58) & 4.90 (3.39) \\ \hline
    6 & \textbf{3.40 (1.98)} & 3.66 (3.50) & 4.38 (3.88) \\ \hline
    Succ. & \textbf{99/100} & 95/100 & 93/100 \\ \hline
    \end{tabular}
\end{table}

\begin{table}[h]
    \centering
    \footnotesize
    \caption{\texttt{077\_rubiks\_cube}}
    \label{tab:077_rubiks_cube}
    \begin{tabular}{|c|c|c|c|}
    \hline
    \multicolumn{4}{|c|}{\texttt{077\_rubiks\_cube}} \\ \hline
    n & DDIM & SDF & SDF (FT) \\ \hline
    1 & \textbf{6.97 (3.68)} & 10.42 (6.93) & 10.85 (7.25) \\ \hline
    3 & \textbf{3.71 (1.59)} & 5.40 (3.23) & 5.43 (2.95) \\ \hline
    5 & \textbf{3.11 (1.31)} & 3.92 (2.44) & 4.15 (2.60) \\ \hline
    6 & \textbf{2.94 (1.47)} & 3.56 (2.22) & 3.69 (2.32) \\ \hline
    Succ. & \textbf{100/100} & 99/100 & \textbf{100/100} \\ \hline
    \end{tabular}
\end{table}

\end{document}